\documentclass[10pt]{article}
\usepackage[letterpaper,textwidth=5.5in,textheight=9in,centering]{geometry}
\usepackage{times}
\usepackage{natbib}
\setcitestyle{authoryear,round,citesep={;},aysep={,},yysep={;}}

\usepackage[T1]{fontenc}
\usepackage[utf8]{inputenc}
\usepackage{microtype}
\usepackage{inconsolata}

\usepackage{latexsym}
\usepackage{wrapfig}
\usepackage{amsmath}
\usepackage{booktabs}
\usepackage{graphicx}
\usepackage{caption}
\usepackage{xcolor}
\usepackage{adjustbox}
\usepackage{multirow}
\usepackage{makecell}
\usepackage{colortbl}
\usepackage{enumitem}
\usepackage[most]{tcolorbox}
\usepackage{listings}
\usepackage{hyperref}
\usepackage{url}

\newcolumntype{C}[1]{>{\centering\arraybackslash}p{#1}}
\newcolumntype{L}[1]{>{\raggedright\arraybackslash}p{#1}}
\title{ARGUS: An Agenda-Driven Framework for Reliable User Simulation}
\date{}

\author{\normalsize Hang Zheng$^{1,2}$\thanks{Hang Zheng and Nan Jiang contribute equally.}, Nan Jiang$^{1,2\ast}$, Zichen Zhu$^{1,2}$, Zixu Yang$^{1,2}$, Zhiyang Tang$^{1,2}$\\
\normalsize Shuai Fan$^{2,3}$, Lu Chen$^{1,2,4}$, Xiaobao Wu$^{1,2}$\thanks{Xiaobao Wu and Kai Yu are the corresponding authors.}, Kai Yu$^{1,2,3,4}$$^{\dag}$ \\
 \normalsize \textsuperscript{1}X-LANCE Lab, School of Computer Science, MoE Key Lab of Artificial Intelligence\\
 \normalsize SJTU AI Institute, Shanghai Jiao Tong University, Shanghai, China\\
 \normalsize \textsuperscript{2}Jiangsu Key Lab of Language Computing, Suzhou, China\\
 \normalsize \textsuperscript{3}AISpeech Co., Ltd., Suzhou, China \quad \textsuperscript{4}Suzhou Laboratory, Suzhou, China\\
\normalsize \texttt{\{azure123, xiaobaowu, kai.yu\}@sjtu.edu.cn}
}

\begin{document}
\maketitle

\begin{abstract}

High-fidelity user simulators are essential for evaluating and training agents in interactive environments. However, current LLM-based simulators struggle with reliability issues, such as goal omission and policy non-compliance. We propose \textsc{Argus} (\textbf{A}genda-d\textbf{R}iven \textbf{G}oal-oriented \textbf{U}ser \textbf{S}imulator), a neuro-symbolic framework that maintains the user's goal state using a dynamic agenda stack manipulated via symbolic operations. By rigorously decoupling state tracking from natural language generation, \textsc{Argus} facilitates strict goal progression and constraint-aligned interactions. For cost-efficient deployment, we distill \textsc{Argus} into compact models using a unified pipeline: evaluator-guided SFT followed by online RL with dense step-level rewards. To rigorously evaluate behavioral fidelity, we introduce a rubric-based metric Composite Behavioral Score (CBS). Evaluations on $\tau^3$-Bench and MultiWOZ demonstrate the significant superiority of the \textsc{Argus} architecture and the efficacy of our training pipeline. Code and data will be released soon.

\end{abstract}

\section{Introduction}
\label{sec:introduction}

As large language models (LLMs) increasingly operate as autonomous agents in open-ended tasks, the demand for robust, scalable interactive environments has surged~\citep{wang2024survey}. Given the prohibitive cost of human-in-the-loop interaction, User Simulators have become indispensable. Beyond its traditional role for automated agent evaluation, a reliable user simulator is now recognized as a critical engine for synthetic data generation across domains like task-oriented dialogue and simulated human feedback~\citep{balog2026usersimulationeragenerative}. Regardless of the application, the core prerequisite is exceptional \textit{behavioral fidelity}: the simulator should strictly adhere to complex, multi-step goals and interaction protocols without hallucination or arbitrary deviation.


Simulating authentic user behavior is inherently a complex sequential decision-making process. Over multiple turns, simulators struggle with maintaining strict \textit{goal progression} (tracking subtask completion to prevent omissions) and enforcing \textit{policy adherence} (following constraints without hallucinating or leaking unprompted information). Existing paradigms manage these challenges suboptimally. Most native LLM-based simulators operate statelessly, collapsing intent reasoning, progress tracking, and response generation into a single forward pass~\citep{terragni2023context}. Despite this flexibility, recent kinetic analyses demonstrate that this heavy cognitive burden inevitably leads to "context drift", which manifests as policy violations and omitted goals in long-horizon tasks~\citep{dongre2025drift}. Conversely, stateful paradigms explicitly track progress by rewriting textual states prior to each turn~\cite{mehri2025goal}. However, framing state transitions as a purely black-box text generation task introduces severe inference overhead and vulnerability to semantic drift, format parsing errors, and temporal confusion.

\begin{figure}[t]
    \centering
    \vspace{-6pt}
    \includegraphics[width=1\linewidth]{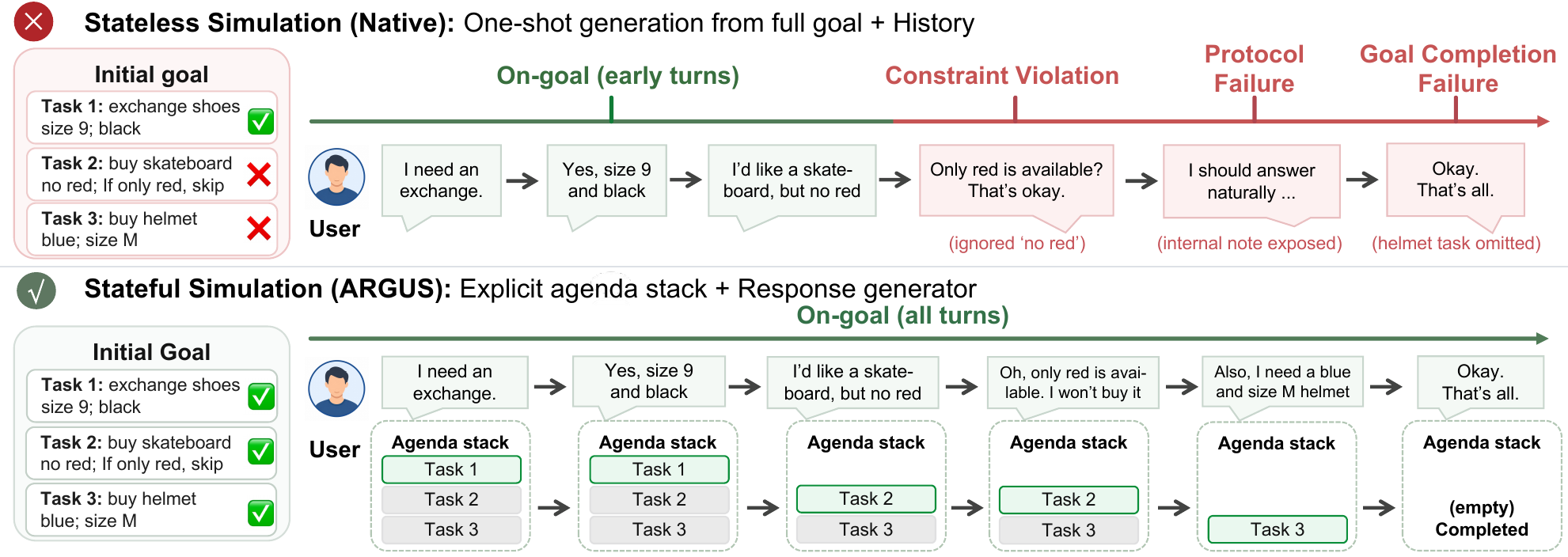}
    \caption{\textbf{Stateless vs.\ stateful user simulation.} Stateless generation may lose constraints or pending goals, while \textsc{Argus} preserves them through an explicit agenda stack.}
    \label{fig:stateful-vs-stateless}
    \vspace{-6pt}
\end{figure}

To address these structural limitations, we propose \textbf{\textsc{Argus}} (\textbf{A}genda-d\textbf{R}iven \textbf{G}oal-oriented \textbf{U}ser \textbf{S}imulator), a neuro-symbolic adaptation of classical agenda-based simulation for LLMs~\citep{schatzmann2007agenda}. Instead of asking an LLM to maintain the entire state through free-form text, \textsc{Argus} maintains a dynamic, Last-In-First-Out agenda stack manipulated via constrained symbolic operations (e.g., \textsc{Push}, \textsc{Pop}). By decoupling state transitions from linguistic realization, \textsc{Argus} offloads discrete logical tracking to the explicit stack while allowing the generation LLM to focus on high-fidelity, contextually grounded language responses. This architectural inductive bias substantially mitigates structural vulnerabilities and enhances rigorous goal progression (illustrated in Figure \ref{fig:stateful-vs-stateless}).

While prompt-based \textsc{Argus} establishes a high behavioral ceiling on frontier models, its two-stage paradigm introduces computational overhead that limits large-scale deployment. To mitigate this cost-quality dilemma, we introduce a two-stage training pipeline to distill the \textsc{Argus} framework into compact models (e.g., 4B/8B). The pipeline uses a shared step-level evaluator to provide behavior-aware supervision for SFT data filtering and online RL reward assignment.
Experimental results show that this pipeline transfers neuro-symbolic tracking to smaller backbones and advances the existing cost-quality Pareto frontier.

\begin{itemize}
    \item We propose \textsc{Argus}, a neuro-symbolic user simulator that uses a LIFO agenda stack to decouple state management from response generation, mitigating goal omission and structural failures in complex interactive tasks.
    \item We design an Evaluator-Guided SFT and RL pipeline that distills high-fidelity simulation into cost-effective 4B/8B models and advances the cost-quality Pareto frontier.
    \item We introduce Composite Behavioral Score, a rubric-based metric that scores behavioral subcriteria and exposes procedural violations missed by outcome-centric metrics.
    \item We demonstrate \textsc{Argus}'s superior behavioral fidelity on $\tau^3$-Bench~\citep{barres2025tau2,cuadron2026saber} and MultiWOZ~\citep{budzianowski2018multiwoz}.

\end{itemize}

\section{Problem Formulation}
\label{sec:formulation}

We formalize user simulation in complex interactive tasks as a sequential decision-making process. The simulator is initialized with a user goal $\mathcal{G}$, comprising constituent tasks, user profiles, and communication policies. At turn $t$, receiving the latest agent message $a_t$ and dialogue history $H_{t-1} = [(a_1, u_1), \ldots, (a_{t-1}, u_{t-1})]$, the simulator generates the next user utterance $u_t$. Existing simulators manage internal control states via two main paradigms.

\begin{figure}[t]
    \centering
    \vspace{-6pt}
    \includegraphics[width=1\linewidth]{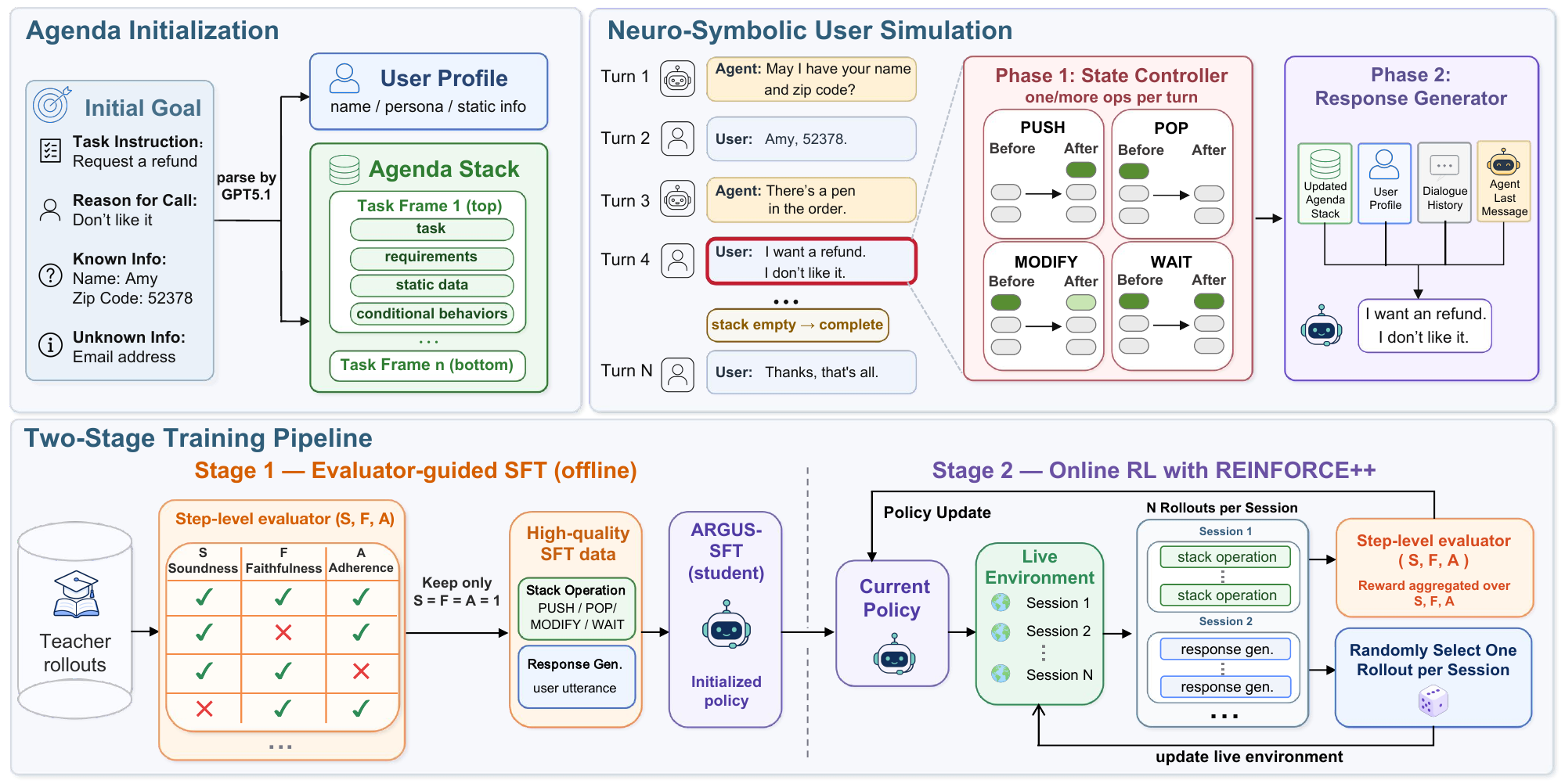}
    \caption{Overview of ARGUS. \textbf{(1) Agenda Initialization:} structured user goals are parsed into a LIFO agenda stack with conditional behaviors. \textbf{(2) Neuro-Symbolic User Simulation:} a state controller updates the stack via symbolic operations (\textsc{Push}, \textsc{Pop}, \textsc{Modify}, \textsc{Wait}), and a response generator produces user utterances. \textbf{(3) Two-Stage Training pipeline:} teacher rollouts filtered by a turn-level evaluator (Soundness, Faithfulness, Adherence) yield SFT data, followed by online RL with step-level rewards.}
    \label{fig:argus-overview}
    \vspace{-6pt}
\end{figure}

\paragraph{Stateless Simulation.} 
Most LLM-based simulators are \textit{stateless}, collapsing goal understanding, intent reasoning, condition verification, and language generation into a single forward pass:
\begin{equation}
u_t \sim \pi_{\theta}(u_t \mid G, H_{t-1}, a_t)
\end{equation}
While conceptually simple, this heavy cognitive burden forces the LLM $\pi_{\theta}$ to implicitly track task progress within its internal representations, often causing structural failures such as state confusion and premature termination as complexity grows. 

\paragraph{Stateful Simulation.} 
To alleviate this, the \textit{stateful} paradigm introduces an explicit runtime state $S_t$, where each user turn is explicitly decoupled into two distinct phases: state transition and response generation. A state transition function $\Phi$ first updates the internal state:
\begin{equation}
    S_{t} = \Phi(S_{t-1}, G, H_{t-1}, a_t)
\end{equation}
Subsequently, the generator produces the final response conditioned on this updated state:
\begin{equation}
    u_t \sim \pi_{\theta}(u_t \mid S_t, G, H_{t-1}, a_t)
\end{equation}

Although recent approaches instantiate $S_t$ as a flat textual dictionary to enhance progress tracking, relying on a purely neural LLM to rewrite these states for deeply nested conditions significantly increases inference overhead and introduces reliability issues, including parsing errors, semantic drift, and temporal confusion.

\section{ARGUS}
\label{sec:methodology}

As established in \S\ref{sec:formulation}, the core challenge of stateful simulation is to design a robust state structure $S_t$ and an effective transition function $\Phi$ that avoids cognitive overload while remaining precise and reliable. To address this, we propose \textbf{\textsc{Argus} (Agenda-dRiven Goal-oriented User Simulator)}. \textsc{Argus} adopts a \emph{neuro-symbolic} architecture, internally maintaining an agenda stack updated through deterministic tool calls. This bypasses the overhead of generating full textual states, but the two-stage design still adds overhead when using large-scale models. We therefore introduce a training pipeline that distills \textsc{Argus} into simulator LLMs with lower latency. Figure~\ref{fig:argus-overview} summarizes the overall system.

\subsection{Neuro-Symbolic User Simulation}
\label{sec:architecture}
\textsc{Argus} processes each user turn in two decoupled phases: state updating and response generation.

\paragraph{State Representation.}
The runtime state $S_t$ comprises a static user profile frame and a dynamic Last-In-First-Out (LIFO) agenda stack. At initialization, a parser (GPT-5.1, Appendix~\ref{sec:appendix_parsing}) converts the textual goal $G$ into a user profile $\mathcal{P}$ (encompassing the persona and communication policies) and a task set $\mathcal{T} = \{T_1, \ldots, T_n\}$, which encapsulates the tasks the simulator must convey and advance, alongside their static attributes and conditional triggers. Each stack frame $f$ is structured as a $\langle \texttt{task}, \texttt{type} \rangle$ pair. We label persistent goals initialized from the structured goal as main tasks and transient interaction needs introduced during dialogue as sub-tasks; both are ordinary agenda frames.

\paragraph{Phase 1: State Updating ($\Phi$).}
The state transition function $\Phi$ leverages the LLM's tool-calling capabilities to execute constrained symbolic operations on the agenda stack:
\begin{equation}
    op_t \in \{\textsc{Push}, \textsc{Pop}, \textsc{Modify}, \textsc{Wait}\}
\end{equation}
At each turn, the state controller emits one or more agenda operations. \textsc{Push} creates a new frame at the top of the stack when an unmet interaction need must be handled before the pending frame. \textsc{Pop} removes the top frame once its corresponding interaction need has been satisfied, subject to the stack lifecycle rules. \textsc{Modify} updates the content or status of a relevant pending frame when an observed agent message triggers a conditional behavior, without changing the stack depth. \textsc{Wait} leaves the agenda unchanged when the latest agent message requires no state transition; the response generator still produces the next user utterance from this unchanged state. Conditioned on the dialogue history and the previous state $S_{t-1}$, the model executes $k$ consecutive operations per turn:
\begin{equation}
    S_t = \left( op_t^{(k)} \circ op_t^{(k-1)} \circ \dots \circ op_t^{(1)} \right)(S_{t-1})
\end{equation}
By confining state updates to executable symbolic operations, \textsc{Argus} reduces the semantic drift and parsing errors common in free-form state rewriting.
\paragraph{Phase 2: Response Generation ($\pi_{\theta}$).}
Conditioned on the updated state $S_t$, the generation LLM $\pi_{\theta}$ produces the natural-language response. This strict \emph{functional decoupling} assigns explicit state-transition decisions to Phase 1, reducing the generator's cognitive burden. The generator receives the full agenda so that lower-priority goals remain visible and can be preserved, while the top frame provides the default immediate focus; lower frames are not advanced unless exposed by valid state operations. Consequently, the model can focus on realizing a response consistent with the updated agenda and dialogue context.

\subsection{Training Pipeline}
\label{sec:training}

While prompt-based \textsc{Argus} excels with frontier LLMs, practical deployment requires cost-efficient inference. We propose a two-stage training pipeline to distill this architecture into compact models (e.g., 4B/8B parameters). A unified \textbf{Step-Level Evaluator} serves as a shared supervision mechanism for SFT data filtering and RL reward assignment.

\paragraph{The Step-Level Evaluator.}
Relying solely on trajectory-level task success to train user simulators has several structural limitations. First, it reduces data efficiency by discarding valid intermediate behaviors from challenging or ultimately unsuccessful dialogues. Second, it may reward behavioral ``shortcuts,'' whereby the simulator violates user constraints or deviates from the intended agenda but nevertheless reaches a successful terminal state. To encourage realistic and consistent user behavior, we employ an LLM-based evaluator that assesses every user step (defined as either Phase 1 or Phase 2 of a turn) via scoring across three dimensions: \emph{Soundness}, \emph{Faithfulness}, and \emph{Adherence} (Appendix~\S\ref{sec:appendix_evaluator}). Each dimension is further operationalized through multiple binary atomic checks, enabling fine-grained and interpretable supervision.


\paragraph{Stage 1: Evaluator-Guided SFT.}
\label{sec:sft}
A strong teacher LLM interacts with a capable agent to generate candidate trajectories. Instead of traditional trajectory-level rejection sampling, we apply our Step-Level Evaluator for rigorous filtering: a user step is retained for supervised fine-tuning (SFT) \emph{only if} it achieves a perfect score across all three dimensions. This discards flawed intermediary behaviors and establishes a robust behavioral prior for the student model.

\paragraph{Stage 2: Online RL with Adapted DAPO.}
\label{sec:rl}

To explore broader interaction distributions and recovery strategies, we further optimize the model through online reinforcement learning using an adapted variant of DAPO~\citep{yu2025dapoopensourcellmreinforcement}. At each training step, we discard rollout groups whose candidates receive identical rewards and continue collecting informative groups until the predefined batch size is reached, thereby ensuring sufficient gradient signal for policy updates. We additionally balance the numbers of Stack and NLG groups within each training batch and maintain equal samples across domains, preventing the training distribution from becoming biased toward particular stages or domains due to differences in task difficulty. 

We use the Step-Level Evaluator as the reward model. During online interaction with the environment, we collect $N$ rollouts for each user step under the same environment state. The evaluator scores each rollout using a fine-grained reward computed as the harmonic mean of the three dimension scores, where each dimension score is the arithmetic mean of its binary atomic checks. These rewards are subsequently used for policy optimization. To advance the live environment, we randomly sample one of the $N$ rollouts rather than greedily selecting the highest-scoring branch. In the spirit of TreeRL~\citep{hou2025treerl}, this strategy encourages broader exploration and reduces overfitting to the reward model. The resulting step-level rewards provide localized behavioral supervision for reducing hallucinations and improving agenda adherence throughout training.
\section{Experimental Setup}
\label{sec:experiments}

\begin{table}[t]
    \centering
    \vspace{-6pt}
    \caption{
        \textbf{CBS (\%) and TSR (\%) results on Retail, Airline and MultiWOZ datasets.}
    }
    \label{table:retail_multiwoz_results}
    \renewcommand{\arraystretch}{1}
    \begin{adjustbox}{width=\textwidth}
    \begin{tabular}{>{\centering\arraybackslash}p{2.5cm}>{\centering\arraybackslash}p{1cm}>{\centering\arraybackslash}p{1cm}|>{\centering\arraybackslash}p{1cm}>{\centering\arraybackslash}p{1cm}|>{\centering\arraybackslash}p{1.2cm}>{\centering\arraybackslash}p{1.2cm}|>{\centering\arraybackslash}p{1.2cm}>{\centering\arraybackslash}p{1.2cm}|>{\centering\arraybackslash}p{1.2cm}>{\centering\arraybackslash}p{1.2cm}|>{\centering\arraybackslash}p{1.2cm}>{\centering\arraybackslash}p{1.2cm}}
        \toprule
        \multirow{2.5}{*}{\makecell{\textbf{Method}}}
        & \multicolumn{2}{c|}{\textbf{Qwen3-8B}}
        & \multicolumn{2}{c|}{\textbf{Qwen3-32B}}
        & \multicolumn{2}{c|}{\textbf{Qwen3-MoE-30B}}
        & \multicolumn{2}{c|}{\textbf{Qwen3-MoE-235B}}
        & \multicolumn{2}{c|}{\textbf{Mistral-Small-24B}}
        & \multicolumn{2}{c}{\textbf{Gemma3-27B}}
        \\
        \cmidrule(lr){2-3}\cmidrule(lr){4-5}\cmidrule(lr){6-7}\cmidrule(lr){8-9}\cmidrule(lr){10-11}\cmidrule(lr){12-13}
        & \textbf{CBS} & \textbf{TSR}
        & \textbf{CBS} & \textbf{TSR}
        & \textbf{CBS} & \textbf{TSR}
        & \textbf{CBS} & \textbf{TSR}
        & \textbf{CBS} & \textbf{TSR}
        & \textbf{CBS} & \textbf{TSR}
        \\
        \midrule

        \rowcolor{gray!8}
        \multicolumn{13}{c}{\textit{Retail}} \\
        \midrule

        Vanilla
        & 50.3 & 44.4
        & 70.7 & 56.0
        & 65.1 & 44.6
        & 73.1 & 50.4
        & 76.6 & 47.5
        & 68.1 & 32.6
        \\

        React
        & 59.6 & 49.6
        & 66.7 & 54.9
        & 56.5 & 45.3
        & 70.9 & 58.9
        & 76.8 & 47.5
        & 75.2 & 39.5
        \\

        Verify
        & 53.3 & 45.3
        & 72.2 & 49.6
        & 63.3 & 44.2
        & 76.8 & 55.4
        & 77.7 & 43.7
        & 69.0 & 38.6
        \\

        Reflection
        & 49.3 & 46.0
        & 69.3 & 53.0
        & 60.2 & 45.4
        & 70.5 & 52.8
        & 70.9 & 44.0
        & 66.9 & 38.6
        \\

        UGST
        & 65.8 & 45.1
        & 63.6 & 54.0
        & 60.7 & 52.3
        & 63.1 & 44.6
        & 68.9 & 49.0
        & 74.7 & 41.8
        \\

        ARGUS (Ours)
        & \textbf{72.5} & \textbf{51.6}
        & \textbf{83.9} & \textbf{57.5}
        & \textbf{85.6} & \textbf{61.1}
        & \textbf{91.3} & \textbf{64.4}
        & \textbf{84.9} & \textbf{61.4}
        & \textbf{85.6} & \textbf{53.8}
        \\

        \midrule

        \rowcolor{gray!8}
        \multicolumn{13}{c}{\textit{Airline}} 
        \\
        \midrule
        
        Vanilla
        & 59.8 & 40.4
        & 64.1 & \textbf{35.6}
        & 62.6 & 42.0
        & 75.1 & \textbf{47.2}
        & 75.3 & \textbf{44.4}
        & 72.2 & 45.2
        \\
        
        React
        & 57.7 & 35.6
        & 59.6 & 32.4
        & 57.7 & 41.6
        & 80.6 & 37.6
        & 73.7 & 41.2
        & 71.4 & 45.2
        \\
        
        Verify
        & 62.9 & 39.6
        & 68.9 & 31.6
        & 60.2 & 42.0
        & 73.3 & 40.4
        & 76.9 & 43.6
        & 74.8 & \textbf{49.2}
        \\
        
        Reflection
        & 63.0 & 42.8
        & 65.8 & 31.2
        & 61.9 & 41.2
        & 76.2 & 40.4
        & 73.4 & 37.6
        & 71.5 & 45.2
        \\
        
        UGST
        & 60.3 & \textbf{44.4}
        & 60.3 & 35.2
        & 62.4 & \textbf{44.4}
        & 72.0 & 41.2
        & 70.9 & 29.6
        & 73.1 & 28.0
        \\
        
        ARGUS (Ours)
        & \textbf{69.0} & 42.4
        & \textbf{78.9} & 34.4
        & \textbf{82.6} & 42.4
        & \textbf{87.0} & 42.4
        & \textbf{77.6} & 39.2
        & \textbf{80.9} & 40.4
        \\

        \midrule

        \rowcolor{gray!8}
        \multicolumn{13}{c}{\textit{MultiWOZ}} \\
        \midrule

        Vanilla
        & 76.5 & 56.8
        & 82.5 & 57.8
        & 83.1 & 54.2
        & 88.4 & 63.2
        & \textbf{82.2} & 47.2
        & 83.4 & 52.7
        \\

        React
        & 74.3 & \textbf{65.4}
        & 72.9 & \textbf{63.4}
        & 76.7 & 57.5
        & 87.5 & 63.5
        & 79.1 & 51.7
        & 76.1 & \textbf{58.6}
        \\

        Verify
        & \textbf{76.7} & 58.6
        & \textbf{83.6} & 59.5
        & 83.1 & 55.2
        & 89.4 & 63.9
        & 81.4 & 44.7
        & 82.3 & 57.3
        \\

        Reflection
        & 75.8 & 57.3
        & 81.1 & 61.0
        & 81.3 & 55.2
        & 87.9 & 64.4
        & 76.7 & 47.9
        & 82.0 & 55.0
        \\

        UGST
        & 67.0 & 50.3
        & 72.0 & 58.8
        & 77.2 & 58.2
        & 82.8 & 64.8
        & 67.9 & 54.6
        & 69.0 & 55.5
        \\

        ARGUS (Ours)
        & 76.0 & 57.4
        & 80.4 & 58.6
        & \textbf{85.2} & \textbf{61.5}
        & \textbf{91.5} & \textbf{67.2}
        & 81.7 & \textbf{65.4}
        & \textbf{84.4} & 57.0
        \\

        \bottomrule
    \end{tabular}
    \end{adjustbox}
    \vspace{-6pt}
\end{table}

\paragraph{Benchmarks.} 
We evaluate simulators on the widely adopted task-oriented dialogue benchmark $\tau^3$-Bench, a corrected version of $\tau^2$-Bench \citep{barres2025tau2,cuadron2026saber}, using two complex executable environments: \textbf{Retail} and \textbf{Airline}. We further adapt \textbf{MultiWOZ} \citep{budzianowski2018multiwoz} into a more challenging interactive environment by normalizing its database into writable domain states, adding transactional APIs, and defining final-state checks. To populate MultiWOZ and construct Retail training sets, we use an automated grounding pipeline that proposes high-level intents, resolves concrete entities and state-dependent values, and derives executable gold solutions and final-state checks. The MultiWOZ test set is additionally manually verified for rigorous evaluation. Retail and MultiWOZ are used for in-domain evaluation, while Airline is held out for OOD evaluation. 
We also show that \textsc{Argus} achieves strong performance with user-side tool use on $\tau^3$-Bench \textbf{Telecom}, and report prompt-based results in Appendix~\S\ref{sec:appendix-telecom-results}. Details of environment conversion and data construction are provided in Appendices~\S\ref{sec:appendix-details} and~\S\ref{sec:appendix-data-construction}.

\paragraph{Baselines \& Simulator Models.} 
We compare \textsc{Argus} against four \textit{stateless} baselines: \textbf{(1) Vanilla}, the native user simulator from $\tau^2$-Bench \citep{barres2025tau2}, \textbf{(2) React}, \textbf{(3) Verify}, and \textbf{(4) Reflection} \citep{yao2024tau}---and one \textit{stateful} baseline, \textbf{(5) UGST} \citep{mehri2025goal}, which maintains an explicit goal-state tracker and conditions user responses on the tracked progress. We evaluate these paradigms across a broad spectrum of frontier LLMs: the Qwen3 series (ranging from 8B to 235B, all instruct versions) \citep{yang2025qwen3technicalreport}, Mistral-Small-24B \citep{mistralai2025mistralsmall3}, and Gemma-3-27B \citep{gemmateam2025gemma3technicalreport}. For the trained simulators, we fine-tune Qwen3 (4B/8B) into five variants: Naive-SFT, UGST-SFT, UGST-RL, \textsc{Argus}-SFT, and \textsc{Argus}-RL. All evaluations are conducted under identical task configurations, with Qwen3-MoE-235B fixed as both the interacting agent and the CBS judger. For the Step-Level Evaluator in ARGUS training, we use GPT-5.1 as the evaluator model. We run each task five times to reduce sampling variance. Detailed baseline implementations and training hyperparameters are provided in Appendix~\S\ref{sec:appendix-implementation}.

\subsection{Evaluation Metrics}
\label{sec:metrics}

\textbf{Task Success Rate (TSR)} is a widely adopted metric in complex interactive tasks that measures whether the agent completes all assigned goals. TSR indirectly reflects the performance of a user simulator because the simulator must communicate its objectives effectively for the agent to succeed. However, it remains fundamentally an \emph{agent-side} signal and has structural limitations as a primary measure of simulator quality. Specifically, TSR is highly susceptible to unrealistic simulator ``shortcuts,'' such as prematurely revealing constraints or hallucinating details, which easily guide the agent to success while violating realistic user protocols. Consequently, TSR serves solely as an auxiliary reference in this work; we provide a detailed discussion regarding its unreliability in \S\ref{sec:illusion-of-success}.

To rigorously evaluate simulator fidelity, we propose the \textbf{Composite Behavioral Score (CBS)} as our primary metric. CBS leverages an LLM-as-a-judge (Qwen3-MoE-235B) to evaluate the dialogue across four rigorous dimensions: \emph{Faithfulness \& Consistency} ($S_{\text{faith}}$), \emph{Policy Adherence} ($S_{\text{policy}}$), \emph{Goal Completion \& Progression} ($S_{\text{goal}}$), and \emph{Format \& Protocol} ($S_{\text{format}}$). For a given dimension to pass, all of its constituent binary sub-criteria must be satisfied. The final CBS is computed as the harmonic mean of these four pass rates:
\begin{equation}
\text{CBS} = \frac{4}{\frac{1}{S_{\text{faith}}} + \frac{1}{S_{\text{policy}}} + \frac{1}{S_{\text{goal}}} + \frac{1}{S_{\text{format}}}}
\label{eq:cbs}
\end{equation}
The harmonic mean penalizes simulators that fail critically in any single aspect. Extensive expert validation confirms that CBS aligns strongly with human judgment and frontier LLMs like GPT-5.4, establishing it as a highly reliable behavioral metric (detailed in Appendix~\S\ref{sec:appendix-cra}).

\section{Empirical Results}
\label{sec:results}

\subsection{Prompt-Based Simulators}
\label{sec:results-prompt}

Table~\ref{table:retail_multiwoz_results} demonstrates that the prompt-based \textsc{Argus} consistently outperforms the baselines across most backbones and domains. This performance gap widens in the more challenging Retail domain, validating the necessity of robust state management for user simulation in complex tasks. Concurrently, Task Success Rates (TSR) generally improve alongside CBS, aligning with our expectations: by strictly adhering to user goals and committing fewer behavioral errors, \textsc{Argus} avoids artificially hindering the agent's task resolution. Controlled variants with matched structured goals and behavioral prompt constraints improve the stateless baselines in several settings, but \textsc{Argus} retains the highest CBS throughout the controlled comparisons (Appendix~\S\ref{sec:appendix-ablation}). We provide a fine-grained analysis of error patterns in ~\S\ref{sec:error-pattern}, highlighting the distinct advantages of \textsc{Argus}.

Interestingly, the performance margin of \textsc{Argus} exhibits an inverse U-shape relative to underlying model capability. For smaller 8B models, the margin is narrow because Phase 1 relies heavily on tool-calling. These models struggle to execute consecutive agenda operations flawlessly, which leads to state preservation errors. Conversely, massive frontier models (e.g., 235B) raise the baseline floor, enabling even Vanilla to achieve high CBS scores. Consequently, the intermediate capability tier is where \textsc{Argus} yields the largest benefits. Nevertheless, \textsc{Argus} retains a clear advantage even in the high-capability regime, demonstrating robust generalizability and structural superiority.

\subsection{Trained Simulators}
\label{sec:results-trained}

\begin{wraptable}[18]{r}{0.52\textwidth}
\vspace{-18pt}
    \centering
    \caption{
        \textbf{CBS (\%) and TSR (\%) for trained simulators.} Airline is the held-out OOD environment. Unless marked, all rows use Qwen3-8B.
        Prompt-based rows are included for comparisons.
    }
\vspace{-3pt}
    \label{table:sft_results}
    \setlength{\tabcolsep}{4pt}
    \resizebox{\linewidth}{!}{%
    \begin{tabular}{lcc|cc||cc}
        \toprule
        \multirow{2}{*}{\makecell{\textbf{Method}}}
        & \multicolumn{2}{c|}{\textbf{Retail}}
        & \multicolumn{2}{c||}{\textbf{MultiWOZ}}
        & \multicolumn{2}{c}{\textbf{Airline}}
        \\
        \cmidrule(lr){2-3}\cmidrule(lr){4-5}\cmidrule(lr){6-7}
        & \textbf{CBS} & \textbf{TSR}
        & \textbf{CBS} & \textbf{TSR}
        & \textbf{CBS} & \textbf{TSR}
        \\
        \midrule
        \rowcolor{gray!8}
        \multicolumn{7}{c}{\textit{Prompt-based}} \\
        \midrule
        Vanilla (8B)
        & 50.3 & 44.4
        & 76.5 & 56.8
        & 59.8 & 40.4
        \\
        ARGUS (8B)
        & 72.5 & 51.6
        & 76.0 & 57.4
        & 69.0 & 42.4
        \\
        Vanilla (235B)
        & 73.1 & 50.4
        & 88.4 & 63.2
        & 75.1 & 47.2
        \\
        ARGUS (32B)
        & 83.9 & 57.5
        & 80.4 & 58.6
        & 78.9 & 34.4
        \\
        \midrule
        \rowcolor{gray!8}
        \multicolumn{7}{c}{\textit{Trained}} \\
        \midrule
        Naive-SFT
        & 62.3 & 47.0
        & 84.5 & 64.0
        & 68.5 & 35.6
        \\
        UGST-SFT
        & 54.3 & 48.6
        & 78.2 & 67.1
        & 57.7 & 38.2
        \\
        UGST-RL
        & 62.0 & 50.7
        & 78.9 & \textbf{69.7}
        & 66.5 & \textbf{39.6}
        \\
        ARGUS-SFT
        & 83.3 & 53.7
        & 85.2 & 61.2
        & 75.9 & 32.0
        \\
        ARGUS-RL
        & \textbf{85.8} & \textbf{58.6}
        & \textbf{87.7} & 65.4
        & \textbf{80.7} & 37.2
        \\
        \bottomrule
    \end{tabular}
    }
\end{wraptable}

To enable cost-efficient deployment, we distill \textsc{Argus} into compact models via our two-stage pipeline. Table~\ref{table:sft_results} compares the 8B trained simulators against prompt-based references (4B results show similar trends; Appendix~\S\ref{sec:appendix-trained-3b}). Trained \textsc{Argus} variants consistently outperform their baseline counterparts. Remarkably, \textsc{Argus}-RL (8B) approaches the performance of the 235B Vanilla and 32B prompt-based \textsc{Argus} simulators, enabling reliable simulation at a fraction of the inference cost (\S\ref{sec:cost-analysis}). The RL stage further improves the trained simulator beyond SFT alone (e.g., increasing Retail CBS from 83.3\% to 85.8\%). As analyzed in \S\ref{sec:sft-vs-rl}, behavior-aware online optimization complements trajectory imitation, particularly on faithfulness and policy adherence.

Furthermore, while UGST-SFT maintains a seemingly competitive TSR on MultiWOZ, its CBS lags significantly. This discrepancy exemplifies the phenomenon of ``illusory success'' (further analyzed in \S\ref{sec:illusion-of-success}). Through case studies, we observe that UGST-SFT achieves a high TSR via unrealistic shortcuts, such as hallucinating task conditions. The substantial behavioral gap between UGST-SFT and \textsc{Argus}-SFT highlights the value of explicit agenda-based state management for preserving goals and constraints during training, while evaluator-guided filtering provides complementary supervision for locally consistent behavior.

\paragraph{Zero-Shot OOD Generalization.}
On the held-out Airline domain, trained \textsc{Argus} shows robust zero-shot out-of-distribution (OOD) generalization. Its strong performance on an unseen domain indicates that the model acquires generalizable, domain-agnostic capability for goal compliance, rather than overfitting to in-domain distributions.

\subsection{Robustness Across Task Complexity}
\label{sec:complexity}

To test whether \textsc{Argus} maintains its advantage as goal complexity increases, we partition the 114 Retail tasks into four balanced tiers---Easy, Medium, Hard, and Expert---by the number of key database-modifying actions and conditional behaviors (Appendix~\ref{sec:appendix-complexity}).

As Figure~\ref{fig:difficulty-curves} illustrates, TSR decreases monotonically with rising complexity, validating our partition. Across all tiers, \textsc{Argus} consistently achieves the highest CBS, demonstrating its robust structural advantage regardless of goal complexity. Synthesizing both metrics reveals the intricate interplay between TSR and CBS. On Easy tasks, TSR saturates due to sufficient agent capability, thereby failing to differentiate simulator quality. Conversely, on Expert tasks, the agent itself becomes the bottleneck; thus, a substantial lead in simulation quality (e.g., \textsc{Argus}'s +10.3\% CBS over Vanilla) yields only a marginal TSR gain (+2.8\%). Only on Medium and Hard tasks does superior simulation prominently translate into higher downstream TSR. This dynamic reinforces the necessity of adopting CBS as the primary behavioral benchmark, rather than relying solely on the task success rate.

\subsection{Inference Cost Analysis}
\label{sec:cost-analysis}

\begin{figure}[t]
  \begin{minipage}[t]{0.485\linewidth}
    \centering
    \includegraphics[width=\linewidth]{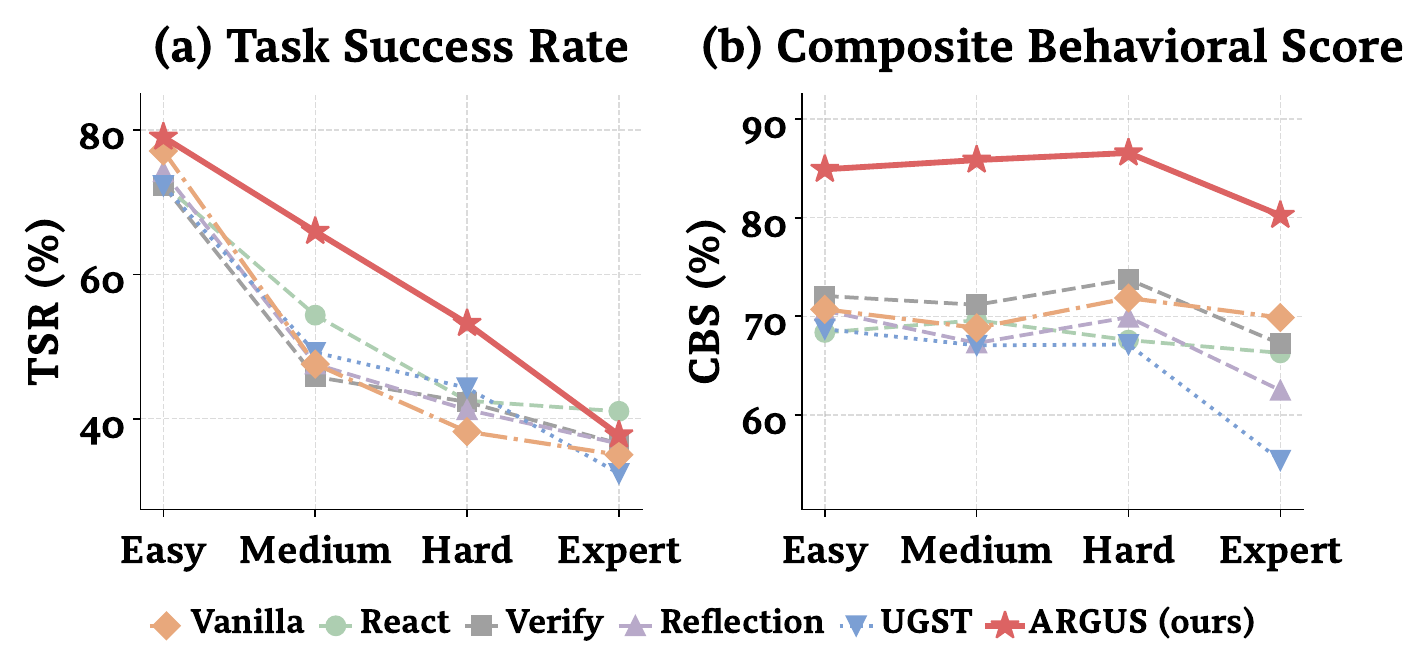}
    \caption{\textbf{TSR and CBS across task complexity levels on Retail.} The curves plot the averaged results across the Qwen model family.}
    \label{fig:difficulty-curves}
  \end{minipage}\hfill
  \begin{minipage}[t]{0.485\linewidth}
    \centering
    \includegraphics[width=\linewidth]{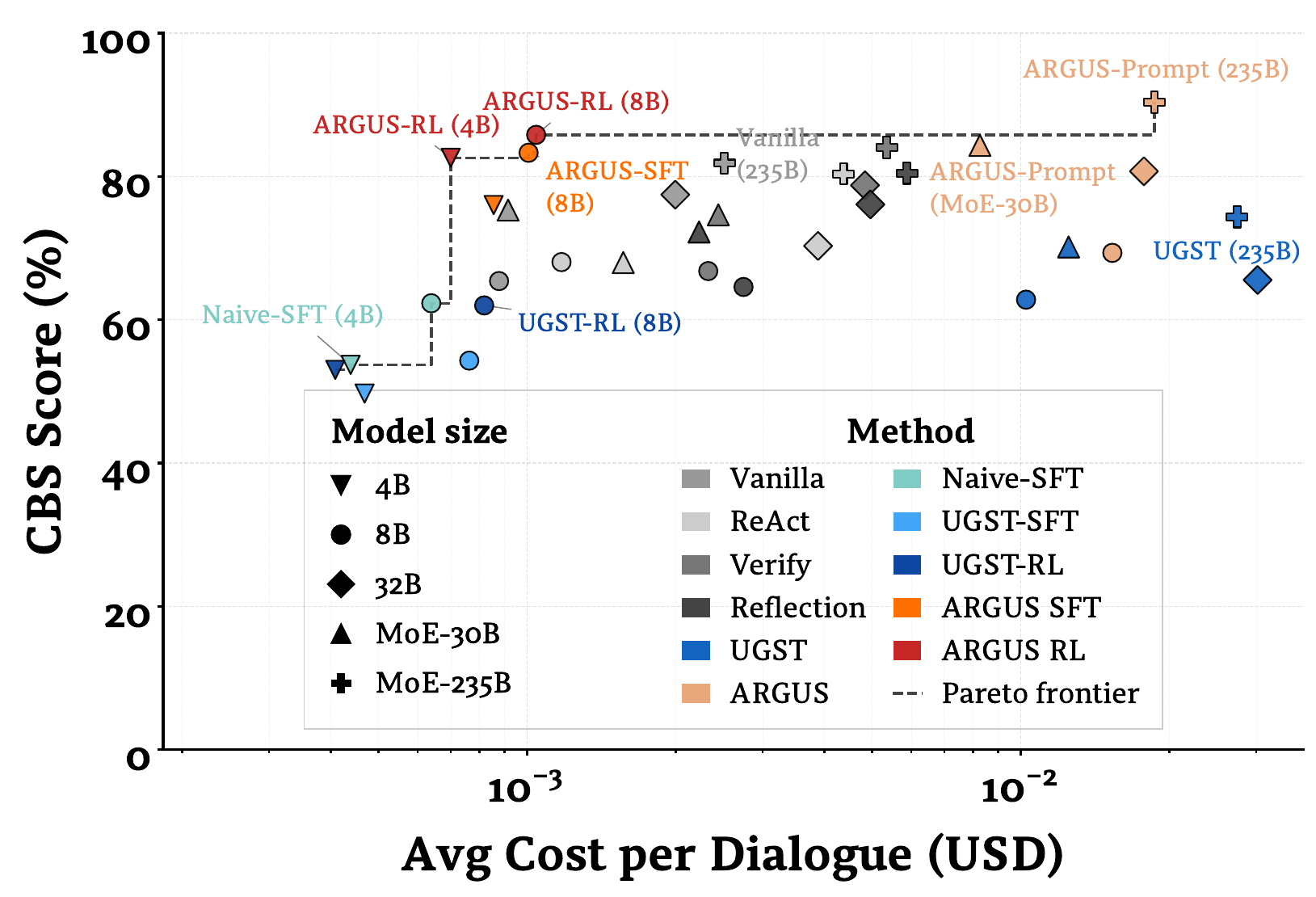}
    \caption{\textbf{Cost--quality trade-off in the Retail domain.} The trained \textsc{Argus} variants advance the existing Pareto frontier, achieving higher CBS than most of the prompt-based simulators at a substantially lower cost.}
    \label{fig:cost-pareto}
  \end{minipage}
\end{figure}

While prompt-based \textsc{Argus} sets a high behavioral ceiling, its two-stage paradigm adds inference overhead, motivating our training pipeline. Figure~\ref{fig:cost-pareto} plots inference cost per dialogue (from official API pricing; Appendix~\S\ref{sec:appendix-cost}) against CBS, with the upper-left region denoting the low-cost, high-quality Pareto frontier. Prompt-based baselines cluster in either low-quality/cheap or high-quality/expensive regions, whereas trained \textsc{Argus} variants advance this frontier. By matching massive 235B configurations at a fraction of the cost, trained \textsc{Argus} mitigates the cost-quality dilemma and enables reliable deployment at large scale.

\section{Analytical Insights}
\label{sec:analysis}

\subsection{Fine-Grained Error Pattern Analysis}
\label{sec:error-pattern}

\begin{figure}[t]
  \centering
  \includegraphics[width=\linewidth]{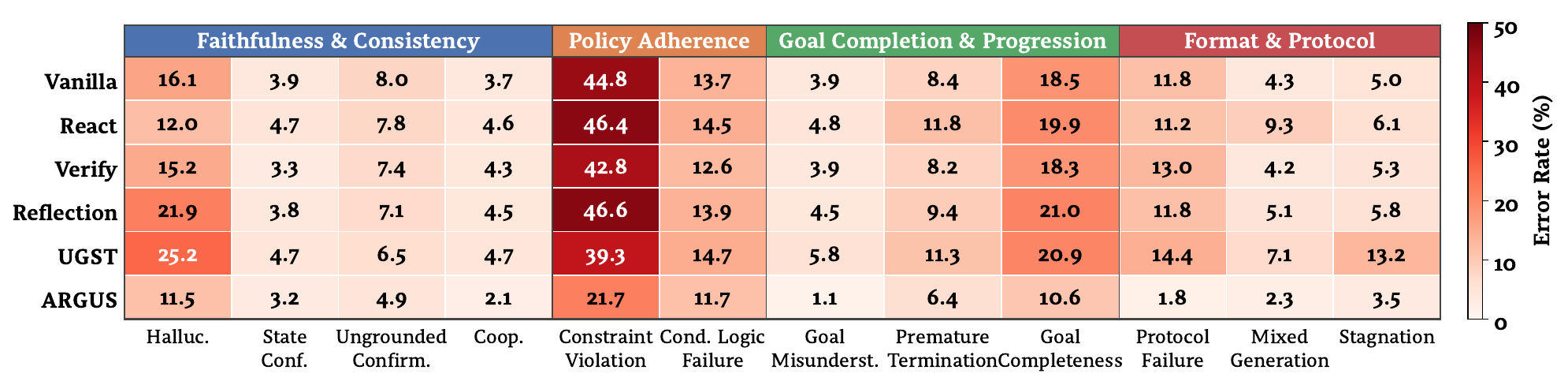}
  \caption{\textbf{Error rate (\%) across 12 CBS subcriteria for six simulators on Retail.} Subcriteria are grouped by their parent dimension. Darker cells indicate higher error rates.}
  \label{fig:error-heatmap}
\end{figure}

Figure~\ref{fig:error-heatmap} details the error rates across 12 CBS subcriteria in the Retail domain. \textsc{Argus} keeps structural errors particularly low: Goal Misunderstanding, Protocol Failure, and Mixed Generation fall to 1.1\%, 1.8\%, and 2.3\%, respectively, compared with 3.9--14.4\% for the baselines. It also substantially reduces premature termination and goal incompleteness, demonstrating the efficacy of functional decoupling and the explicit LIFO stack in preventing format contamination and tracking pending subgoals. Although Constraint Violation remains a shared bottleneck, \textsc{Argus} reduces its error rate to 21.7\%, compared with 39.3--46.6\% for the baselines. The smaller differences on State Confusion (3.2\% vs. 3.3--4.7\%) and Conditional Logic Failure (11.7\% vs. 12.6--14.7\%) indicate that the gains of \textsc{Argus} are concentrated in structural control, particularly format compliance and state tracking, rather than reflecting a universal improvement in reasoning.

\subsection{The Illusion of Success}
\label{sec:illusion-of-success}

Our fine-grained error analysis reveals a critical measurement flaw: baseline simulators often violate behavioral protocols while still guiding the agent to task success. We define an \emph{Illusory Success} (False Positive, FP) as a dialogue where the agent succeeds ($\text{TSR} = 1$) despite simulator violations ($\text{CBS} < 1$). The False Positive Rate ($\text{FPR} = \frac{\text{FP}}{\text{TP} + \text{FP}}$) measures the share of reported successes built on flawed simulations.

As Figure~\ref{fig:quadrant} illustrates, baselines exhibit severe FPRs of 48.1\%--60.6\%, whereas \textsc{Argus} reduces the FPR to 33.6\%. Qualitative analysis indicates that many ``illusory successes'' arise from unrealistic simulator shortcuts, including the premature disclosure of key information or future trajectories and unconditional compromise when the agent struggles. These behaviors artificially ease the agent's decision-making burden. This also suggests that for RL-based agent training, using TSR alone could therefore induce reward hacking, encouraging agents to exploit simulator loopholes rather than improve genuine task-solving ability. For benchmarking, isolating dialogues where the simulator fully satisfies all criteria (CBS=1, covering both Genuine Successes [TP] and Unfair Failures [FN]) yields a stricter testbed that better reveals agent capability boundaries. Appendix~\S\ref{sec:appendix-false-sr} provides further analysis and a concrete shortcut case study.

\subsection{Advantage of RL over SFT}
\label{sec:sft-vs-rl}

\begin{figure}[t]
  \begin{minipage}[t]{0.485\linewidth}
    \centering
    \includegraphics[width=\linewidth]{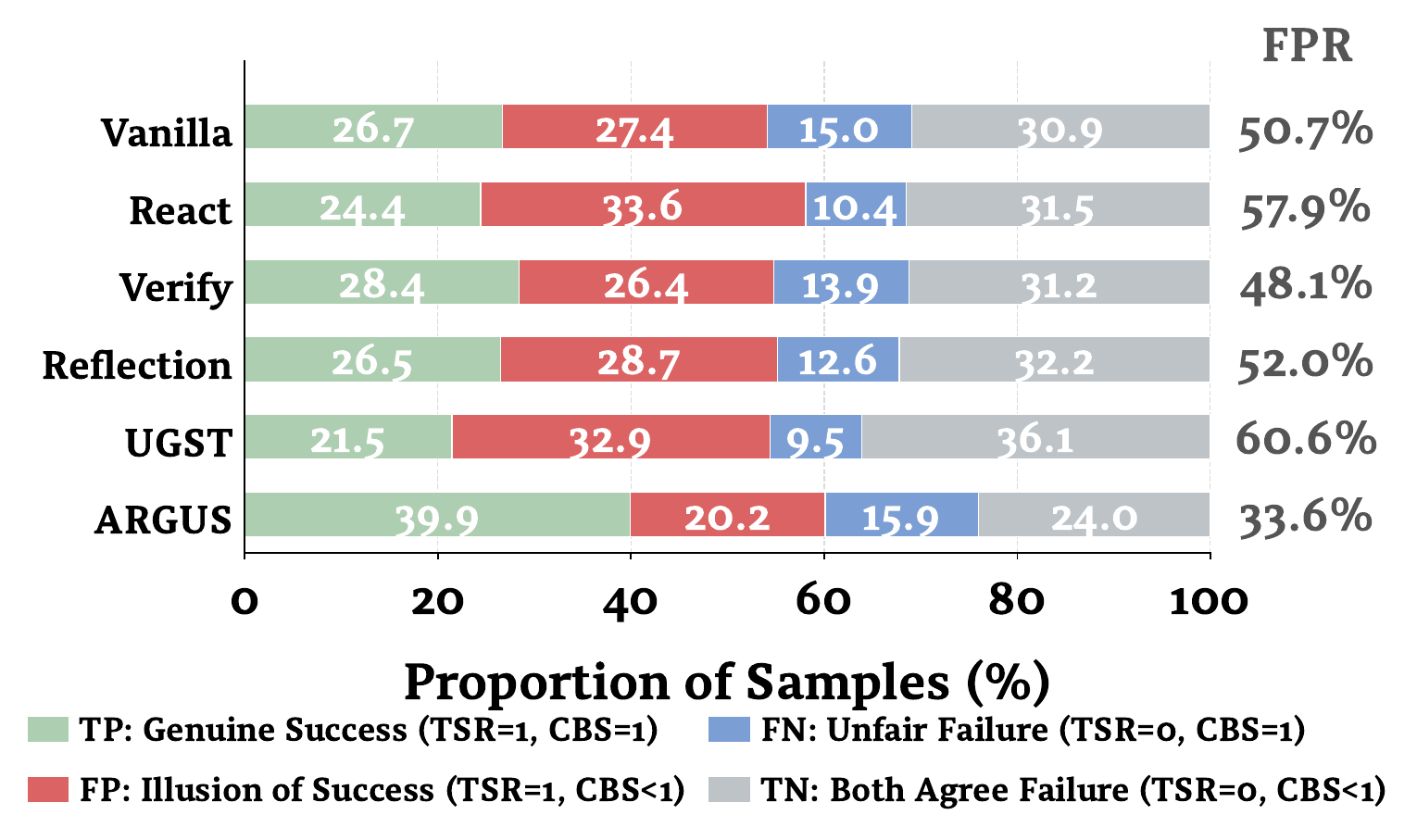}
    \caption{\textbf{Four-quadrant breakdown by TSR outcome and CBS quality, aggregated across Retail and MultiWOZ using four Qwen3 backbones. }The red segment (\emph{Illusory Success}) represents dialogues where TSR reports success despite underlying simulator violations.}
    \label{fig:quadrant}
  \end{minipage}\hfill
  \begin{minipage}[t]{0.485\linewidth}
    \centering
    \includegraphics[width=\linewidth]{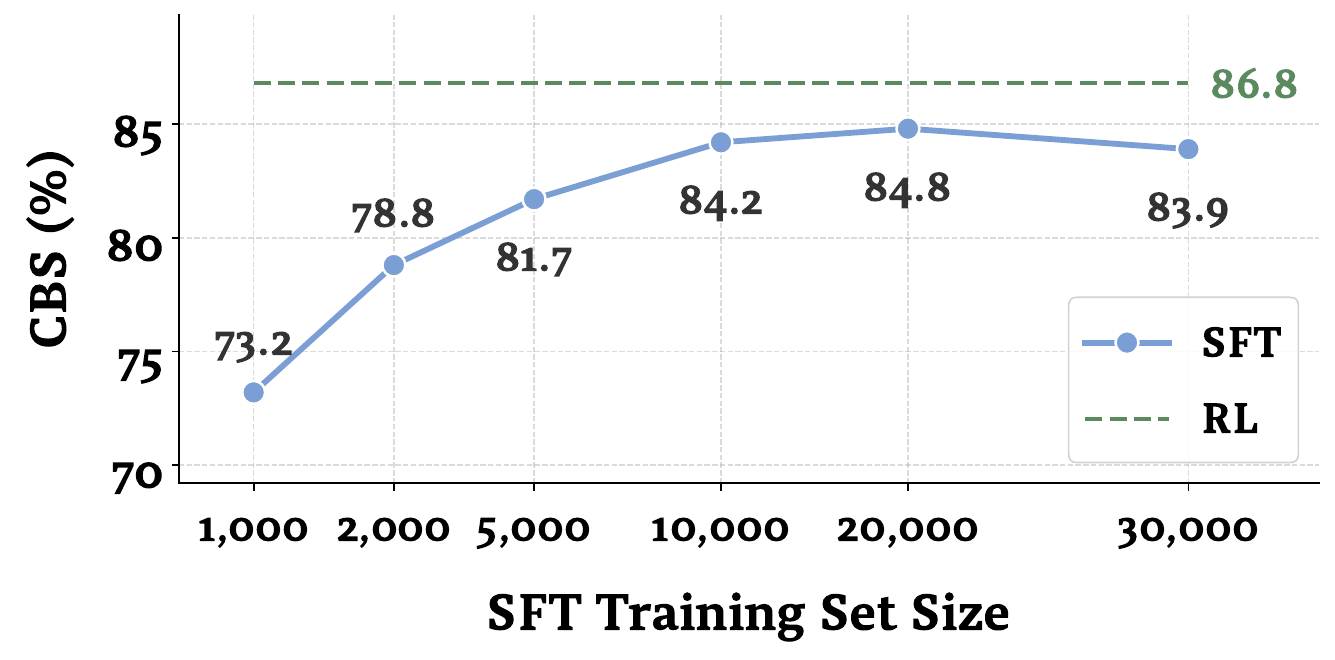}
    \caption{\textbf{CBS of \textsc{Argus}-SFT across training data sizes
    (1k--30k), compared with \textsc{Argus}-RL (8B).}
    Scores are averaged across Retail and MultiWOZ.}
    \label{fig:sft-vs-rl}
  \end{minipage}
\end{figure}

A natural question is whether RL is necessary, or whether more SFT data suffices. Figure~\ref{fig:sft-vs-rl} plots CBS against SFT data volume: performance improves up to 20k samples, then plateaus and slightly regresses at 30k, never surpassing RL. A dimensional breakdown (Appendix~\S\ref{sec:appendix-sft-vs-rl}) shows that RL outperforms SFT in \emph{Faithfulness \& Consistency} and \emph{Policy Adherence}, suggesting that behavior-aware online optimization complements passive trajectory imitation, particularly for rule adherence and hallucination avoidance.

\section{Related Work}
\label{sec:related-work}

\paragraph{Agenda-Based User Simulators.} Classical user simulators use explicit goals and agenda-based dialogue acts to support controllable state updates and policy learning \citep{schatzmann2007agenda,schatzmann2009hidden}.

\paragraph{LLM-Based User Simulators.}

LLM-based simulators replace hand-crafted acts with prompting or in-context learning, making user simulation more flexible across domains \citep{terragni2023context,davidson2023user,chen2021teaching,algherairy2025prompting}. Related work also uses LLMs for interactive evaluation, data generation, fine-tuning, and verification \citep{kazi2024large,cheng2022multiwoz,niu2024enhancing,sekulic2024reliable,luo2024duetsim}, while another line studies diversity, persona control, and implicit user profiles \citep{liu2023one,ahmad2025simulating,nakano2025generating,gromada2025evaluating,wang2025know,sun2023metaphorical,naous2025flipping}.


\paragraph{Stateful User Simulators.}

UGST is the most relevant stateful baseline, tracking user-goal progress and using the tracked state to guide response generation \citep{mehri2025goal}.
Recent methods add persona control or structured reasoning over dynamic goals and trajectories \citep{karthikeyan2025agentic,zhang2026userlmr1modelinghumanreasoning}.
ARGUS differs by making state operational through a LIFO agenda stack updated by constrained symbolic operations, rather than treating state as a descriptive summary or trace.

\section{Conclusion}
\label{sec:conclusion}

This paper presents ARGUS, a novel neuro-symbolic framework that addresses the persistent challenges of goal omission, context drift, and policy non-compliance in LLM-based user simulators. By decoupling logical state tracking via a dynamic LIFO agenda stack from natural language generation, ARGUS enhances rigorous goal progression and high behavioral fidelity. To overcome the inference costs associated with large models, we introduced a unified two-stage distillation pipeline comprising evaluator-guided SFT and behavior-aware online RL. This pipeline transfers these capabilities to compact 4B and 8B models. Evaluations across complex interactive environments, using our newly proposed multi-dimensional Composite Behavioral Score (CBS), demonstrate that ARGUS structurally advances the existing cost-quality Pareto frontier. 

Despite these gains, \textsc{Argus} has two main limitations. Errors in goal parsing or agenda updates may propagate to response generation because the current pipeline does not explicitly verify or revise its intermediate state. Moreover, the agenda-based design assumes that goals can be represented as structured tasks, and its extension to open-ended conversations without clear target states remains to be studied. Ultimately, ARGUS provides a scalable, highly reliable solution for synthetic data generation and the rigorous evaluation of autonomous agents.

\newpage

\subsection*{AI use statement}
Generative AI tools were used to assist with manuscript organization, language
editing, and grammar checking. We also used generative AI models to generate
synthetic task and dialogue data used in this study. The authors reviewed the
AI-assisted writing and manually checked the generated data through the
validation procedures described in the paper and appendices. The authors take
responsibility for the final content of this work, including all claims,
analyses, data, and other artifacts produced with the aid of generative AI.

\subsection*{Ethics statement}
All human annotations in this work were conducted internally by the authors
acting as expert annotators. No external participants or crowdworkers were
recruited, and the evaluated materials consist only of synthetic or
benchmark-derived task descriptions and dialogue transcripts without
personally identifiable information. ARGUS is intended for research on the
evaluation and training of task-oriented dialogue systems. Because simulated
users may reproduce biases in the underlying models and benchmarks or be
misused outside the studied setting, the results should be interpreted within
the task-oriented domains evaluated in this work. Further details of the human
evaluation setup are provided in Appendix~\S\ref{sec:appendix-human-eval-setup}.

\subsection*{Reproducibility statement}
We describe the ARGUS architecture and training procedure in
\S\ref{sec:methodology}, with implementation details and hyperparameters in
Appendix~\S\ref{sec:appendix-implementation}. The appendices further document
structured-goal extraction, data construction and validation, benchmark
composition, the complete CBS evaluation protocol, evaluator reliability
analyses, and the prompts used in our experiments. These materials, together
with the main-paper experimental setup and results, provide the information
needed to reproduce the reported evaluations.

\bibliography{custom}
\bibliographystyle{plainnat}

\appendix
\newtcblisting[auto counter]{promptlisting}[3][]{%
    enhanced,
    breakable,
    listing only,
    width=\columnwidth,
    colback=black!1,
    colframe=black!45,
    coltitle=black,
    colbacktitle=black!9,
    fonttitle=\sffamily\bfseries\footnotesize,
    title={Prompt~\thetcbcounter: #2},
    label={#3},
    boxrule=0.5pt,
    arc=1.2pt,
    outer arc=1.2pt,
    left=6pt,
    right=6pt,
    top=7pt,
    bottom=7pt,
    boxsep=2pt,
    toptitle=2pt,
    bottomtitle=2pt,
    before skip=0.9em plus 0.15em minus 0.1em,
    after skip=1.1em plus 0.15em minus 0.1em,
    listing options={
        basicstyle=\ttfamily\footnotesize,
        breaklines=true,
        breakatwhitespace=false,
        columns=fullflexible,
        keepspaces=true,
        showstringspaces=false,
        tabsize=2
    },
    #1
}
\section{Initial Goal Extraction and Agenda Initialization}
\label{sec:appendix_parsing}

This appendix describes the initialization step used in \S~\ref{sec:architecture}. The goal of this component is to convert a natural-language user scenario into a structured goal state, which is then used to initialize the static user profile and the initial agenda stack of \textsc{Argus}.

\subsection{Extraction Pipeline}

For each benchmark instance, the input is the full natural-language user scenario, including the user's background information, private constraints, task instructions, and reason for contacting the agent. This scenario is inserted into the extraction prompt in Prompt~\ref{prompt:goal-extraction}. The model output must be a strict JSON object containing a global user profile and a list of task objects.

The extracted JSON is verified before it is used. We first check whether the output is parseable JSON. If parsing succeeds, a verifier model receives the raw scenario and the generated JSON, and scores whether the JSON faithfully represents the scenario under the schema rules in Prompt~\ref{prompt:goal-verification}. If the verifier returns a failing score, the extraction model revises the JSON using the verifier feedback. We retry this extract-and-verify loop up to five times.

\subsection{Structured Goal State}

The extracted \texttt{user\_profile} stores global persona, communication policies, and static data such as names, zip codes, email addresses, or other hard facts. Each extracted task contains four components: the task itself, a \texttt{requirements} field describing the user's requested outcome and constraints, task-specific \texttt{static\_data}, and optional \texttt{conditional\_behaviors}. The extracted \texttt{tasks} are used to initialize the agenda stack in scenario order: the task that should be handled first is placed at the top of the stack, while later tasks are placed below it. We separate default requirements from conditional behaviors because the stack controller must trigger fallback, narrowing, or abandonment logic only when the corresponding agent message makes the condition true.

\paragraph{Illustrative example.}
The following example shows the complete structured representation of a multi-item Retail request:

\begin{lstlisting}[
  basicstyle=\ttfamily\footnotesize,
  breaklines=true,
  breakatwhitespace=false,
  columns=fullflexible,
  keepspaces=true,
  showstringspaces=false,
  frame=single
]
{
  "user_profile": {
    "name": "Yusuf Rossi",
    "personality": ["Detail-oriented"],
    "communication_policy": [
      "Address all requests in a single turn if possible"
    ],
    "static_data": {
      "zip_code": "19122"
    }
  },
  "tasks": [
    {
      "task": "Modify Order #W2378156 (Keyboard & Thermostat)",
      "requirements": "Exchange the mechanical keyboard for a similar one (must be clicky, have RGB backlight, and be full size). Also exchange the smart thermostat for one compatible with Google Home instead of Apple HomeKit.",
      "static_data": {
        "order_id": "#W2378156"
      },
      "conditional_behaviors": [
        {
          "condition_index": 0,
          "trigger_condition": "The agent reports that no keyboard matches all requirements (clicky, RGB, and full size).",
          "instruction": "Accept a keyboard without a backlight as long as the other requirements match."
        }
      ]
    }
  ]
}
\end{lstlisting}

The user profile is stored globally, while the task object initializes a persistent agenda frame containing both requested exchanges. The conditional behavior remains inactive until the agent reports that no keyboard satisfies all requirements; once triggered, its instruction updates the relevant task while preserving the thermostat request.

\section{Step-Level Evaluator}
\label{sec:appendix_evaluator}

This appendix details the step-level evaluator used in \S~\ref{sec:training}. Each user turn in \textsc{Argus} contains two separately judged steps: the stack operation step and the natural-language generation step. The evaluator first audits whether the stack operations are correct, and then audits whether the final user response is consistent with the updated agenda stack and the original goal. For supervised fine-tuning, a candidate turn is retained only if both the stack operation step and the natural-language generation step pass all three evaluation dimensions. For reinforcement learning, the two stages are evaluated separately and their dimension-level scores provide dense rewards for the corresponding model outputs.

\subsection{Stage 1: Stack Logic Evaluation}

The stack logic evaluator receives the user goal, conversation history, latest agent message, stack before the operation, simulator reasoning, and the executed stack operations. It judges \texttt{push\_task}, \texttt{pop\_task}, \texttt{modify\_top\_task}, and \texttt{wait} under three binary dimensions: \emph{Soundness}, \emph{Faithfulness}, and \emph{Adherence}. Soundness checks stack lifecycle correctness, including timely pushes, non-premature pops, cascade popping of completed subtasks, and main-task popping only after explicit completion. Faithfulness checks that operation arguments and reasons are grounded in the user goal or dialogue context. Adherence checks whether conditional behaviors and persona constraints are triggered exactly when required. The full prompt is shown in Prompt~\ref{prompt:stack-logic}.

\subsection{Stage 2: NLG Consistency Evaluation}

The NLG consistency evaluator receives the updated agenda stack, user profile, dialogue history, the agent's latest message, and the generated user response. It again uses the three binary dimensions. Soundness checks whether the response directly and completely fulfills the updated stack top without scope creep. Faithfulness checks whether all factual content is grounded in the goal, profile, or dialogue. Adherence checks persona, communication policy, privacy, bundling policy, and response realism. The full prompt is shown in Prompt~\ref{prompt:nlg-consistency}.

\paragraph{Alignment with trajectory-level CBS.}
To examine whether the training-time step scores align with trajectory-level behavioral quality, we analyze 300 complete Retail trajectories spanning prompt-based \textsc{Argus} with 8B and 235B backbones and \textsc{Argus}-RL with a 8B backbone. For each trajectory, we average its step-level evaluator scores and group the trajectories into non-empty score buckets. Table~\ref{tab:reward-alignment} shows that the mean step score is strongly associated with mean CBS across the eight bucket averages (Spearman's $\rho=0.9762$, $p<0.001$; Pearson's $r=0.9463$, $p<0.001$). CBS broadly increases with the mean step score, with a small decline in the highest-score bucket. This aggregate analysis indicates that the training-time evaluator tracks the trajectory-level behavioral quality measured by CBS, supporting its use as a behavior-aware supervision signal.

\begin{table}[ht]
\centering
\caption{\textbf{Alignment between trajectory-aggregated step scores and final CBS.} The 300 trajectories are grouped into eight non-empty score buckets; no trajectory falls in $[0, 0.2]$.}
\label{tab:reward-alignment}
\small
\setlength{\tabcolsep}{5pt}
\renewcommand{\arraystretch}{1.0}
\begin{tabular}{@{}lrrr@{}}
\toprule
\textbf{Mean step-score bucket} & \textbf{$N$} & \textbf{Mean step score} & \textbf{Mean CBS (\%)} \\
\midrule
$(0.2, 0.3]$ & 8  & 0.2751 & 27.30 \\
$(0.3, 0.4]$ & 12 & 0.3584 & 46.70 \\
$(0.4, 0.5]$ & 30 & 0.4625 & 62.18 \\
$(0.5, 0.6]$ & 39 & 0.5618 & 70.86 \\
$(0.6, 0.7]$ & 51 & 0.6480 & 73.42 \\
$(0.7, 0.8]$ & 54 & 0.7552 & 82.37 \\
$(0.8, 0.9]$ & 55 & 0.8618 & 90.78 \\
$(0.9, 1.0]$ & 51 & 0.9452 & 87.09 \\
\bottomrule
\end{tabular}
\end{table}

\section{Interactive MultiWOZ Environment}
\label{sec:appendix-details}

We convert MultiWOZ into an executable environment with writable domain state, shared service layers, and final-state verification. Starting from the original MultiWOZ entities and constraints, we add normalized schemas, controllable state, transactional records, contact and payment information, and executable APIs for hotel, restaurant, train, taxi, attraction, and public-service domains, together with shared geographic, contact, payment, and utility services. The resulting MultiWOZ test set contains 155 tasks, including 125 single-domain tasks and 30 multi-domain tasks.

\subsection{Data Cleaning and State Construction}

We first normalize the original database. Entity identifiers are converted to strings and standardized with domain-specific prefixes, such as \texttt{hotel\_}, \texttt{rest\_}, and \texttt{attraction\_}. Postcodes are lowercased and stripped of spaces. Missing or inconsistent key fields, including address, phone number, and location, are repaired when necessary. For public-service domains, we enrich hospital and police records with missing addresses and geographic coordinates. Hospital records are deduplicated and separated into hospital-level metadata and department-level entries.

We introduce a controllable seven-day state window starting from 2026-01-05. Dates are represented in \texttt{YYYY-MM-DD} format and times use 24-hour \texttt{HH:MM} format. The state covers domains that require availability checks or transactional updates, including hotels, restaurants, trains, and taxis. Each domain maintains its own writable inventory and booking state, while time, geography, contact, and payment are represented as shared service layers. This separation allows single-domain tasks to be generated and verified without unintended interference, while still supporting cross-domain dependencies when a task explicitly requires them.

For hotels, we add room catalogs and room types, together with daily prices, inventory, and booking calendars. Hotel state also includes attributes such as breakfast inclusion and checkout time, enabling tasks involving room-type constraints, price comparison, unavailable options, booking modification, and cancellation.

For restaurants, we add table types, structured opening hours, reservation availability, and slot-level table inventory. Reservation state is generated for lunch and dinner periods with 30-minute granularity. This supports tasks involving unavailable time slots, party-size constraints, fallback times, and reservation changes.

For trains, we retain the original train identifiers and add seat inventory over the seven-day state window. Inventory is organized by class and seat type. Seat maps distinguish window and aisle preferences, with A/D seats treated as window seats and B/C seats treated as aisle seats; first and business classes use A/D seats only. Train bookings are written into state, enabling lookup, modification, and cancellation.

For taxis, we generate a fleet of vehicles with capacity, car type, color, reservation support, base fare, and per-kilometer fare. Driver shifts are generated over the state window, and taxi bookings update driver availability, contact records, and payment state.

For attractions, ticketed attractions may support reservations, while attractions without paid entry are treated as non-reservable.

We also introduce a contact and payment layer. Each user is identified by a unique pair of \texttt{contact\_name} and \texttt{contact\_phone}. Contacts maintain balances and booking histories. Paid booking, modification, and cancellation tools validate the paid amount, update balances, and record booking-level payment information. This enables tasks involving insufficient balance, incorrect payment amounts, refunds, and payment-dependent fallback behavior.

\subsection{Tool Design}

The API design follows three principles. First, each transactional domain supports a closed-loop interaction pattern: search, availability or feasibility checking, booking, modification, cancellation, and booking lookup. Second, state-changing tools are verifiable: booking, modification, and cancellation operations update domain state, inventory, calendars, contact records, and payment information when applicable. Third, tools enforce identity and payment constraints. Modification and cancellation require matching contact information, and paid operations validate the required amount before updating state.

Table~\ref{tab:multiwoz-tools} summarizes the supported operations by domain and shared service layer.

\begin{table}[t]
\centering
\caption{\textbf{Supported operations in the interactive MultiWOZ environment.}}
\label{tab:multiwoz-tools}
\setlength{\tabcolsep}{2pt}
\renewcommand{\arraystretch}{1.0}
\begin{tabular}{p{0.18\linewidth}p{0.76\linewidth}}
\toprule
\textbf{Domain or layer} & \textbf{Supported operations} \\
\midrule
Hotel & Search by static attributes and dynamic constraints; check room availability and price; create, modify, cancel, and retrieve bookings. \\
\rowcolor{gray!20}Restaurant & Search by area, food type, price range, and reservation support; check table availability; create, modify, cancel, and retrieve reservations. \\
Train & Search by route, date, time, class, seat preference, and availability; assign seats; create, modify, cancel, and retrieve bookings. \\
\rowcolor{gray!20}Taxi & Search by route, time, capacity, vehicle attributes, and reservation support; estimate fares; create, modify, cancel, and retrieve taxi bookings. \\
Attraction & Search by area, type, name, and reservation support; check reservation feasibility; create, modify, cancel, and retrieve reservations when supported. \\
\rowcolor{gray!20}Geo and utility & Compute distances between places, retrieve nearby entities within a target domain, and perform simple arithmetic calculations. \\
Contact and payment & Validate contact identity, maintain balances and booking histories, process paid operations, and record refunds when applicable. \\
\rowcolor{gray!20}Public service & Search hospitals, departments, and police stations; retrieve detailed public-service information. \\
\bottomrule
\end{tabular}
\end{table}

\subsection{Closed-loop Behavior and Verification}

For each transactional domain, tools support a complete execution loop. Search tools retrieve candidate entities under static and optional dynamic constraints. Availability or search-based feasibility checks verify whether a candidate can satisfy the requested date, time, capacity, room type, seat type, or route constraint. Booking tools create records and update domain state. Modification tools verify contact identity, recompute availability and price differences, and update inventory and payment state. Cancellation tools mark bookings as cancelled, release reserved resources, and refund balances when applicable.

This closed-loop design allows tasks to include realistic failure and fallback paths. For example, a hotel task may require fallback when the requested room type is unavailable; a restaurant task may require selecting another time slot; a train task may require falling back from a preferred seat type; and a taxi task may require comparing fares before booking. Since these tools update state, task success can be evaluated through final-state checks rather than surface-form matching.

\subsection{Cross-domain Support}

The converted environment supports cross-domain task construction. Geographic tools enable constraints such as nearest hotels, restaurants near attractions, taxi routes between selected entities, and maximum-distance requirements. Temporal state enables dependencies among train arrival times, taxi pickup times, restaurant reservations, and hotel check-in dates. The contact and payment layer supports cost-sensitive tasks, including insufficient-balance failures, alternative selections, and refund verification.

These additions make the converted MultiWOZ environment suitable for interactive user simulation and agent evaluation, where success depends on both retrieving correct information and performing valid state-changing actions under realistic constraints.

\section{Data Construction Details}
\label{sec:appendix-data-construction}

\subsection{Case Construction}

We use the following construction pipeline to generate Retail and MultiWOZ training and evaluation cases. The pipeline automatically transforms high-level user intents into executable tasks and verifiable final-state checks, while ensuring that the generated cases are grounded in the actual database and environment state. All construction stages are performed with GPT-5.4.

We first generate a general case description from the environment background, database and state format, available task types, and few-shot examples. The description specifies the user's high-level intent, the intended interaction pattern, and possible failure or fallback rules, but does not yet bind the task to concrete entities or database values.

We then ground the case through an agentic planning process. The planner interacts with the actual database and state through a small set of controlled resolvers, including initial booking construction, pool-based case selection, search, and candidate selection. Search handles filtering and feasibility checks over the database and state, while candidate selection can directly choose one returned search result as the entity or value needed by the case. To keep the process controllable, each planning step calls only one resolver and resolver arguments are expressed by referring to previous results rather than by inventing free-form values.

As part of pool-based case selection, we maintain precomputed pools built from the database and environment state. These pools improve both feasibility and coverage. Common-case pools cover ordinary valid scenarios, while corner-case pools target rare or difficult situations such as inventory boundaries, partially available slots, ambiguous candidates, price extremes, and fallback-triggering failures. During planning, the model may directly select cases from these pools or use them as anchors for grounding more complex cases.

The planner iteratively updates the case according to resolver feedback, such as missing candidates, non-unique results, or confirmed shortfalls. It may revise non-essential details, add grounded constraints, or apply deterministic tie-breaking rules while preserving the core user intent. Cases that cannot be grounded into feasible and coherent tasks are discarded or regenerated.

The output of this stage is an agentic trajectory that records resolver calls, selected entities, grounded facts, state-dependent values, and fallback decisions. Based on this trajectory and the initial case description, we generate the natural-language user task, the simulator initial state, and a gold executable solution used to derive the expected final-state check. The generation follows the trajectory order, so the user task reflects the intended information-disclosure process. Although the gold solution records the intended action sequence, task success is ultimately determined by comparing the resulting environment state against the expected final-state check.

\subsection{Case Validation}

Generated cases are validated through a two-stage process. First, we use a solo-mode procedure from $\tau^2$-Bench \citep{barres2025tau2}. An LLM is given the generated task and asked to solve it directly by executing actions in the environment, without simulating a full user-agent conversation. We then compare the final database or environment state against the expected final-state check. This validation is performed as a single attempt without retry; if the final state satisfies the check on that attempt, the case passes the first validation stage.

Second, we use another LLM as an auditor. The auditor compares the task, trace, and expected final-state check to verify that the task is faithful to the grounded facts, all user constraints are supported by the trace, the expected final state correctly captures task completion, and the case does not contain ambiguity or missing information.

We retain only cases that pass both validation stages, which helps ensure the correctness of the generated tasks and final-state checks. For the MultiWOZ test set, we additionally perform \textbf{manual review} to further ensure case correctness.

\subsection{Human Evaluation Setup}
\label{sec:appendix-human-eval-setup}

All human evaluations in this work were conducted internally by the authors acting as expert annotators. No external participants or crowdworkers were recruited, and therefore no financial compensation was applicable. For MultiWOZ test-set verification, annotators were instructed to check whether each constructed task was grounded in the database state, whether the user requirements were unambiguous and executable, and whether the expected final-state check correctly captured task completion. For CBS alignment validation, annotators followed the CBS criteria and binary subcriteria defined in Appendix~\S\ref{sec:appendix-cbs-protocol}. The evaluated materials consist only of synthetic or benchmark-derived task descriptions and dialogue transcripts, without personally identifiable information.

\section{CBS Evaluation Framework}
\label{sec:appendix-cra}

\subsection{Evaluation Protocol}
\label{sec:appendix-cbs-protocol}

The Composite Behavioral Score (CBS) evaluates simulator quality along four dimensions, each decomposed into binary subcriteria. Table~\ref{tab:cra-subcriteria} lists all 12 subcriteria. For each dialogue, a subcriterion is scored 1 (pass) or 0 (fail). A dimension passes only when all of its subcriteria pass (strict AND logic). The final CBS score is the harmonic mean of the four dimension-level average pass rates across all evaluated dialogues. The full dimension-level evaluator prompts are provided in Prompts~\ref{prompt:cbs-faithfulness-consistency}, \ref{prompt:cbs-policy-adherence}, \ref{prompt:cbs-goal-completion}, and~\ref{prompt:cbs-format-protocol}.

\begin{table}[tbp]
\centering
\caption{\textbf{The 12 binary subcriteria of the CBS evaluation framework, grouped by dimension.}}
\label{tab:cra-subcriteria}
\setlength{\tabcolsep}{2pt}
\renewcommand{\arraystretch}{1.0}
\begin{tabular}{@{}p{0.18\linewidth}p{0.23\linewidth}p{0.53\linewidth}@{}}
\toprule
\textbf{Dimension} & \textbf{Subcriterion} & \textbf{Description} \\
\midrule

\multirow{11}{*}{\makecell[l]{D1: Faithfulness\\\& Consistency}}
  & \multirow{3}{*}{\makecell[l]{Hallucination}}
  & Simulator fabricates information in the response, introducing new details not present in the scenario or conversation facts, such as fictitious order numbers, account information, device status, or other business details. \\
  & \cellcolor{gray!20}\multirow{2.5}{*}{\makecell[l]{State\\Confusion}}
  & \cellcolor{gray!20}Simulator confuses the conversation state, mixing up attributes of different entities, different orders, different slots, or different check results, which leads to incorrect factual correspondences. \\
  & \multirow{2.5}{*}{\makecell[l]{Ungrounded\\Confirm}}
  & Simulator makes ungrounded confirmations, directly saying "yes" or "that's right" without verification when the Agent proposes candidate information that is unprovided or incorrect. \\
  & \cellcolor{gray!20}\multirow{2.5}{*}{\makecell[l]{Cooperativeness}}
  & \cellcolor{gray!20}Simulator fails to cooperate truthfully, baselessly pretending not to know, shifting responsibility, or glossing over information that it already knows in the scenario and should have answered. \\

\midrule
\multirow{6}{*}{\makecell[l]{D2: Policy\\Adherence}}
  & \multirow{3}{*}{\makecell[l]{Constraint\\Violation}}
  & Simulator violates constraints by prematurely leaking information it should not say, exposing meta-instructions, or accepting options and out-of-boundary solutions explicitly prohibited by the scenario. \\
  & \cellcolor{gray!20}\multirow{3}{*}{\makecell[l]{Cond. Logic\\Failure}}
  & \cellcolor{gray!20}Simulator fails to execute conditional logic correctly, meaning the if -> then rules in the scenario are either not triggered when they should be (missed trigger) or triggered when they should not be (false trigger). \\

\midrule
\multirow{8}{*}{\makecell[l]{D3: Goal\\Completion\\\& Progression}}
  & \multirow{3}{*}{\makecell[l]{Goal\\Misunderstanding}}
  & Simulator misunderstands the goal and veers off the task direction, such as attempting to cancel instead of reschedule, attempting to refund instead of exchange, or getting the target object wrong. \\
  & \cellcolor{gray!20}\multirow{3}{*}{\makecell[l]{Premature\\Termination}}
  & \cellcolor{gray!20}Simulator causes omission or premature termination, ending the conversation early while unexpressed task requirements remain, or stopping the interaction before reaching a reasonable end point. \\
  & \multirow{2}{*}{\makecell[l]{Goal Completeness}}
  & Simulator does not omit any tasks it should request and ensures that all of them have been executed. \\

\midrule
\multirow{9}{*}{\makecell[l]{D4: Format\\\& Protocol}}
  & \cellcolor{gray!20}\multirow{2.5}{*}{\makecell[l]{Protocol\\Failure}}
  & \cellcolor{gray!20}Simulator violates the protocol format by failing to output the precise STOP termination token as required at the end of the conversation, either by omitting or misspelling it. \\
  & \multirow{3}{*}{\makecell[l]{Mixed\\Generation}}
  & Simulator mixes unintended content into the generation, blending natural language with structured commands/special tokens in the same turn to disrupt the protocol, or confusing the user/agent roles by speaking in the agent's tone. \\
  & \cellcolor{gray!20}\multirow{3}{*}{\makecell[l]{Stagnation}}
  & \cellcolor{gray!20}Simulator causes the conversation to stall through mechanical repetition or generation lock-ups, such as continuously repeating the exact same sentence or ignoring the Agent's new inputs without providing new information. \\

\bottomrule
\end{tabular}
\end{table}

The judge model (Qwen3-MoE-235B) receives the full dialogue transcript together with the original task specification and evaluates each subcriterion independently. To reduce variance, we use structured prompting with explicit pass/fail definitions for each subcriterion. The strict AND aggregation within dimensions ensures that a single violation in any subcriterion causes the entire dimension to fail, reflecting the principle that a reliable simulator must avoid \emph{all} categories of errors rather than compensating one type with strength in another.

The harmonic mean makes CBS sensitive to the weakest behavioral dimension rather than rewarding high average performance alone. For example, a simulator scoring 0.95 on three dimensions but only 0.35 on the fourth receives a CBS of 0.665, below a simulator that scores 0.80 on all four dimensions. This reflects our evaluation goal: a reliable simulator should avoid critical failures across all CBS dimensions, rather than compensating for a severe weakness in one dimension with strong performance in others.

\subsection{Human Alignment Validation}
\label{sec:appendix-evaluator-reliability}

We validate the reliability of CBS in two ways: first, by checking whether expert annotators can apply the rubric consistently, and second, by measuring whether LLM judges agree with human labels and with each other. This validation is intended to test the rubric itself, not only the particular judge model used in our main experiments.

We randomly sample 90 dialogues, with 45 from Retail and 45 from MultiWOZ, covering 3 simulator types and 3 backbone models. Five authors of this paper act as expert annotators and label the 12 binary CBS subcriteria. To estimate inter-annotator agreement (IAA), all annotators label the same 15 shared cases. The remaining 75 cases are split across annotators so that each annotator labels 30 cases in total, including the 15 shared cases and 15 unique cases. This design yields 150 human--LLM annotation pairs for each subcriterion. We then evaluate Qwen3-MoE-235B, our primary CBS judge, against the human labels, and run GPT-5.4 with the same evaluator prompts as an independent cross-model check.

Because CBS labels are highly imbalanced, with most subcriteria passing in most dialogues, we report Gwet's AC2 for the multi-annotator IAA setting and Gwet's AC1 for pairwise comparisons. These metrics are better suited to high-prevalence binary labels than agreement measures whose chance correction is dominated by marginal label frequencies \citep{gwet2008computing}.

Table~\ref{tab:cbs-reliability} consolidates the validation results. Human annotators reach strong overall agreement (AC2 = 0.861), indicating that the CBS rubric is sufficiently clear for consistent human application. Qwen3-MoE-235B also aligns closely with human labels (AC1 = 0.812), and GPT-5.4 provides a consistent independent check (human agreement AC1 = 0.748; Qwen--GPT agreement AC1 = 0.775). Agreement is strongest on objective structural subcriteria such as Protocol Failure, State Confusion, and Stagnation, and weakest on Conditional Logic Failure, where judging whether a conditional rule was correctly enforced is inherently more context-dependent. We therefore use Qwen3-MoE-235B as the primary judge in the main experiments, while the GPT-5.4 results provide additional evidence that the CBS rubric is not tied to a single judge model.

To further test whether CBS-based model comparisons depend on the choice of
judge, we conduct a full-cell robustness analysis on \textsc{Retail} using
DeepSeek-V4.1-Flash as an alternative judge. This analysis covers 24 method
cells formed by four Qwen-series backbones and six user-simulation methods.
As shown in Table~\ref{tab:retail-cbs-judge-agreement}, the cell-level CBS
scores produced by Qwen3-MoE-235B and DeepSeek are strongly correlated
(Pearson's $r=0.873$, Spearman's $\rho=0.886$, and Kendall's
$\tau_b=0.746$). Agreement reaches $93.81\%$ with Cohen's $\kappa=0.624$
for Protocol Failure. More importantly, the two judges agree on $91.67\%$
of within-backbone method orderings, with an aggregate-ranking Kendall's
$\tau_b=0.966$. Under task-level cluster bootstrap resampling, the
probability that ARGUS ranks first under the DeepSeek judge ranges from
$95.49\%$ to $99.92\%$ across the four backbones. These results indicate
that the relative method rankings, particularly the leading performance
of ARGUS, are robust to the choice of CBS judge.

\begin{table}[ht]
    \centering
    \small
    \caption{\textbf{Agreement between the Qwen and DeepSeek CBS judges on Retail domain.} Cell-level correlations are computed over 24 method cells.}
    \label{tab:retail-cbs-judge-agreement}
    \begin{tabular}{lc}
        \toprule
        Metric & Agreement \\
        \midrule
        Cell CBS Pearson $r$ & 0.873 \\
        Cell CBS Spearman $\rho$ & 0.886 \\
        Cell CBS Kendall $\tau_b$ & 0.746 \\
        Protocol Failure agreement & 93.81\% \\
        Protocol Failure Cohen's $\kappa$ & 0.624 \\
        Method-pair ranking agreement & 91.67\% \\
        Aggregate-ranking Kendall $\tau_b$ & 0.966 \\
        DeepSeek $P(\text{ARGUS ranks first})$ & 95.49--99.92\% \\
        \bottomrule
    \end{tabular}
\end{table}

\begin{table}[t]
\centering
\caption{\textbf{Consolidated CBS reliability results across human inter-annotator agreement, human--LLM agreement, and Qwen--GPT model agreement. }Human IAA uses Gwet's AC2 because it is a multi-rater setting; pairwise comparisons use Gwet's AC1.}
\label{tab:cbs-reliability}
\setlength{\tabcolsep}{0.5pt}
\renewcommand{\arraystretch}{1.0}
\begin{tabular*}{\textwidth}{@{\extracolsep{\fill}}llcccc@{}}
\toprule
\multirow{2.5}{*}{\textbf{Dimension}} & \multirow{2.5}{*}{\textbf{CBS Subcriterion}}
& \multicolumn{1}{c}{\textbf{Human IAA}}
& \multicolumn{1}{c}{\textbf{Human--Qwen}}
& \multicolumn{1}{c}{\textbf{Human--GPT}}
& \multicolumn{1}{c}{\textbf{Qwen--GPT}} \\
\cmidrule(lr){3-3}\cmidrule(lr){4-4}\cmidrule(lr){5-5}\cmidrule(lr){6-6}
& & AC2 & AC1 & AC1 & AC1 \\
\midrule
\multirow{4}{*}{\makecell[l]{Faithfulness\\\& Consistency}}
& Hallucination         & 0.896 & 0.807 & 0.528 & 0.705 \\
& State Confusion       & 1.000 & 0.846 & 1.000 & 0.901 \\
& Ungrounded Confirm.   & 0.928 & 0.935 & 0.539 & 0.638 \\
& Cooperativeness       & 0.939 & 0.929 & 0.916 & 0.887 \\
\midrule
\multirow{2}{*}{Policy Adherence}
& Constraint Violation  & 0.828 & 0.603 & 0.701 & 0.739 \\
& Cond.\ Logic Failure  & 0.464 & 0.498 & 0.518 & 0.413 \\
\midrule
\multirow{3}{*}{\makecell[l]{Goal Completion\\\& Progression}}
& Goal Misunderstanding & 0.765 & 0.850 & 0.773 & 0.862 \\
& Premature Termination & 0.739 & 0.744 & 0.589 & 0.814 \\
& Goal Completeness     & 0.740 & 0.652 & 0.623 & 0.556 \\
\midrule
\multirow{3}{*}{\makecell[l]{Format\\\& Protocol}}
& Protocol Failure      & 0.973 & 0.973 & 0.986 & 0.977 \\
& Mixed Generation      & 0.966 & 0.835 & 0.562 & 0.689 \\
& Stagnation            & 0.906 & 0.907 & 0.963 & 0.912 \\
\midrule
\multicolumn{2}{l}{\textbf{Overall}}
& \textbf{0.861} & \textbf{0.812} & \textbf{0.748} & \textbf{0.775} \\
\bottomrule
\end{tabular*}
\end{table}

\section{Implementation Details}
\label{sec:appendix-implementation}

\subsection{ARGUS Implementations}
\label{sec:appendix-argus-implement}

As a decoupled two-phase architecture, the core state management of ARGUS is driven by an external agenda stack, which we implement as a dedicated Python class that securely maintains the dynamic Last-In-First-Out (LIFO) structure.

\paragraph{Prompt-Based Implementation} 
Phase 1 and 2 utilize distinct system prompts (Prompts \ref{prompt:argus-phase1} and \ref{prompt:argus-phase2}). Because we found that open-weight LLMs (e.g., Qwen series) struggle to simultaneously reason and execute tool calls in a single turn, Phase 1 provides five OpenAI-format tools: an explicit \texttt{think} tool, alongside four core stack operations (\textsc{Push}, \textsc{Pop}, \textsc{Modify}, \textsc{Wait}). The LLM must invoke \texttt{think} before updating the stack. Consecutive tool calls are parsed and applied sequentially to the Python class. To ensure stack safety, any invalid or malformed tool calls are automatically intercepted and ignored, preventing execution. In the current implementation, we do not utilize an exception-handling mechanism (e.g., returning error messages to prompt a retry loop), which remains a direction for future optimization. In Phase 2, the model is not provided with any external tools and instead strictly utilizes a ReAct-style reasoning format to generate the final natural language response based on the updated stack.

\paragraph{Trained Implementation} 
For the trained ARGUS variants, the complex two-phase operational logic is deeply internalized into the model weights through our training pipeline. Having learned the syntax and semantics of the stack operations natively, they no longer require the heavy tool-calling workarounds or the extensive instruction sets used in the prompt-based setup. Consequently, the system prompts for the trained models (detailed in Prompts \ref{prompt:argus-trained-phase1} and \ref{prompt:argus-trained-phase2}) are significantly more concise, allowing the distilled models to manipulate the stack and generate high-fidelity responses with minimal input overhead.

\subsection{Baseline Implementations}
\label{sec:appendix-baselines}

We provide implementation details for the baseline simulators used in our experiments. ReAct, Verify, and Reflection are prompt-only baselines adapted from $\tau$-Bench-style user simulation \citep{yao2024tau}. They receive the same user-simulation guidelines, task scenario, and dialogue history, and do not call user-side tools or maintain an explicit external state. UGST is a stateful baseline for goal-aligned user simulation \citep{mehri2025goal}.

\textbf{ReAct.}
Following the $\tau$-Bench prompting baseline \citep{yao2024tau}, ReAct adds a private reasoning step before the visible user response. The model emits a \texttt{<thought>} block and a separate \texttt{<response>} block; only the response block is returned to the environment when parsing succeeds. This tests whether explicit private reasoning improves simulator behavior without adding a persistent state representation.

\textbf{Verify.}
Following the $\tau$-Bench verifier baseline \citep{yao2024tau}, Verify performs rejection sampling with an LLM verifier. After generating a candidate user response, the model is prompted as a binary verifier with temperature set to zero. If the verifier accepts the candidate, the candidate is returned; otherwise, generation is retried up to three times.

\textbf{Reflection.}
Following the $\tau$-Bench reflection baseline \citep{yao2024tau}, Reflection extends Verify with a self-correction step. If the initial candidate is rejected, the LLM is prompted to write a reflection and a revised user response. The revised response is then verified again before being returned.

\textbf{UGST.}
UGST is based on user-goal state tracking for goal-aligned user simulation \citep{mehri2025goal}. It initializes a structured user-goal state from the task, with fields for user profile, user policy, task objectives, requirements, and preferences. At each turn, a state-tracking LLM updates these fields using deterministic decoding before the response LLM generates the next user utterance conditioned on the textual scenario, dialogue history, latest agent message, and updated state. Thus, unlike the prompting-only baselines, UGST explicitly tracks goal progress and constraint alignment across the conversation.

The exact prompts used for the three prompting-only baselines are shown in Prompts~\ref{prompt:baseline-react}, \ref{prompt:baseline-verify}, and~\ref{prompt:baseline-reflection}.

\subsection{Training Details}
All training experiments are conducted using the veRL framework~\citep{sheng2024hybridflow} on eight NVIDIA A800 GPUs.

\paragraph{Supervised Fine-Tuning (SFT).}
All SFT experiments (Naive-SFT, UGST-SFT, and \textsc{Argus}-SFT) employ the following hyperparameters: 10,000 training samples, a maximum sequence length of 8192, a batch size of 64, a learning rate of $1 \times 10^{-5}$, and 1 training epoch. For Naive-SFT and UGST-SFT, the training data is constructed by extracting teacher-simulator trajectories (Qwen3-MoE-235B) that successfully complete the task, without additional step-level filtering. In contrast, \textsc{Argus}-SFT utilizes the Evaluator-Guided filtering mechanism to rigorously select training samples step-by-step, as detailed in \S\ref{sec:training}.

\paragraph{Reinforcement Learning (RL).}
All RL experiments (UGST-RL and \textsc{Argus}-RL) utilize the following hyperparameters: a batch size of 64, a KL penalty coefficient of 0.005, $N=4$ rollouts per sample, and a total of 25 training steps. UGST-RL uses pre-sampled interaction states for rollout collection and step scoring. For \textsc{Argus}-RL, we customize the data collection module within the veRL framework to dynamically acquire rollout data and corresponding reward signals directly from the live $\tau^3$-Bench environment, thereby strictly preserving the on-policy nature of the training process. To optimize efficiency, we implement a parallel session manager for the $\tau^3$-Bench environment that maintains 24 concurrent interaction sessions. This asynchronous design provides a robustness margin, ensuring that the buffer accumulates a full batch of training data within approximately a single interaction round per step, thereby substantially reducing the overall training time. 

\section {Supplementary experimental results}
\subsection{Additional Prompt-Based Results}
\label{sec:appendix-telecom-results}

Table~\ref{tab:telecom-prompt-results} reports supplementary prompt-based results on the Telecom environment, where user simulation requires tool-use interaction. These results provide additional evidence that \textsc{Argus} remains effective in a harder user-simulation setting; we place the full table in the appendix due to space limits.
\begin{table}[t]
\centering
\caption{\textbf{Prompt-based simulator results on the Telecom environment, which requires tool-using interaction.}}
\label{tab:telecom-prompt-results}
\small
\setlength{\tabcolsep}{4pt}
\renewcommand{\arraystretch}{1.05}
\begin{adjustbox}{width=\textwidth}
\begin{tabular}{lcc|cc|cc|cc|cc|cc}
\toprule
\multirow{2}{*}{\textbf{Method}}
& \multicolumn{2}{c|}{\textbf{Qwen3-8B}}
& \multicolumn{2}{c|}{\textbf{Qwen3-32B}}
& \multicolumn{2}{c|}{\textbf{Qwen3-MoE-30B}}
& \multicolumn{2}{c|}{\textbf{Qwen3-MoE-235B}}
& \multicolumn{2}{c|}{\textbf{Mistral-Small-24B}}
& \multicolumn{2}{c}{\textbf{Gemma3-27B}} \\
\cmidrule(lr){2-3}\cmidrule(lr){4-5}\cmidrule(lr){6-7}
\cmidrule(lr){8-9}\cmidrule(lr){10-11}\cmidrule(lr){12-13}
& \textbf{CBS} & \textbf{TSR}
& \textbf{CBS} & \textbf{TSR}
& \textbf{CBS} & \textbf{TSR}
& \textbf{CBS} & \textbf{TSR}
& \textbf{CBS} & \textbf{TSR}
& \textbf{CBS} & \textbf{TSR} \\
\midrule
Vanilla
& 53.2 & 19.7
& 64.2 & 24.7
& 51.2 & 28.1
& 71.9 & 34.7
& 48.2 & 19.8
& 50.5 & 18.8 \\
React
& 57.8 & 21.1
& 65.1 & 25.1
& 54.9 & 27.0
& 73.9 & 47.5
& 49.7 & 27.4
& 51.5 & 20.5 \\
Verify
& 56.7 & 19.5
& 64.7 & 22.5
& 51.8 & 29.7
& 68.8 & 33.9
& 53.6 & 22.5
& 58.1 & 17.0 \\
Reflection
& 47.3 & 18.7
& 63.4 & 24.9
& 53.5 & 31.9
& 70.0 & 36.0
& 60.6 & 27.3
& 62.3 & 17.4 \\
UGST
& 49.2 & 17.4
& 49.4 & 19.0
& 45.0 & 21.1
& 66.6 & 38.8
& 47.7 & 32.8
& 48.7 & \textbf{27.0} \\
\textsc{ARGUS} (Ours)
& \textbf{62.5} & \textbf{22.3}
& \textbf{69.7} & \textbf{29.5}
& \textbf{68.1} & \textbf{34.9}
& \textbf{80.0} & \textbf{48.6}
& \textbf{69.3} & \textbf{38.6}
& \textbf{71.9} & 26.5 \\
\bottomrule
\end{tabular}
\end{adjustbox}
\end{table}

\subsection{Ablation Study}
\label{sec:appendix-ablation}
We ablate the structured \textsc{Argus} action space and the step-level evaluator by introducing them incrementally. The trajectory-based variant retains samples only from dialogues that ultimately achieve task success, whereas the step-level variant applies the evaluator described in Appendix~\S\ref{sec:appendix_evaluator} to retain only locally correct user steps. As shown in Table~\ref{tab:filtering-ablation}, the step-level variants generally achieve higher CBS scores, indicating more faithful user simulation, while the trajectory-based variants tend to obtain relatively higher TSR. To better understand this discrepancy, we further inspect cases where the step-level evaluator rejects a sample but the trajectory-based criterion accepts it. We observe shortcut behaviors similar to those discussed in Appendix~\S\ref{sec:appendix-shortcut-cases}, where the simulator reduces task difficulty by prematurely providing or leaking information.

More importantly, the results suggest that the sample-filtering strategy is not the primary factor distinguishing the performance of different methods. Both \textsc{Argus}-SFT variants achieve substantially higher CBS scores than their Naive-SFT and UGST-SFT counterparts. This indicates that \textbf{the main advantage of \textsc{Argus} comes from its neuro-symbolic architecture}: by decomposing simulation into two stages and explicitly maintaining and updating a state stack, \textsc{Argus} achieves markedly higher simulation reliability.

\begin{table}[t]
    \centering
    \caption{\textbf{Ablation of architecture and SFT data filtering.} All variants use Qwen3-8B. ``Trajectory'' keeps samples only from task-successful trajectories, while ``step-level'' applies the evaluator in Appendix~\S\ref{sec:appendix_evaluator}. For Naive-SFT and UGST-SFT, we design step evaluators tailored to their respective requirements.}
    \label{tab:filtering-ablation}
    \setlength{\tabcolsep}{4pt}
    \begin{tabular}{l cc cc}
        \toprule
        \multirow{2.5}{*}{\textbf{Method}}
        & \multicolumn{2}{c}{\textbf{Retail}}
        & \multicolumn{2}{c}{\textbf{MultiWOZ}} \\
        \cmidrule(lr){2-3}\cmidrule(lr){4-5}
        & \textbf{CBS} & \textbf{TSR}
        & \textbf{CBS} & \textbf{TSR} \\
        \midrule
        Naive-SFT (trajectory) & 62.3 & 47.0 & 84.5 & 64.0 \\
        Naive-SFT (step-level) & 66.8 & 43.8 & 84.6 & 65.7 \\
        UGST-SFT (trajectory) & 54.3 & 48.6 & 78.2 & \textbf{67.1} \\
        UGST-SFT (step-level) & 62.6 & 47.2 & 81.4 & 64.9 \\
        \textsc{Argus}-SFT (trajectory) & 81.9 & 51.9 & 84.5 & 63.6 \\
        \textsc{Argus}-SFT (step-level) & \textbf{83.3} & \textbf{53.7} & \textbf{85.2} & 61.2 \\
        \bottomrule
    \end{tabular}

\end{table}

\paragraph{Input-representation and prompt controls.}
To test whether the prompt-based gains arise from the parsed goal representation or more detailed behavioral instructions, we augment Vanilla and ReAct with the structured goal used by \textsc{Argus}, matched prompt constraints, or both. Table~\ref{tab:prompt-fairness-controls} reports both CBS and TSR. The controls improve the stateless baselines in several settings, confirming that representation and prompt design matter, but their effects vary across backbones and domains and are not consistently additive. \textsc{Argus} nevertheless achieves the highest CBS in every tested setting. TSR follows a less consistent pattern, reinforcing its role as an auxiliary outcome metric rather than a direct measure of simulator fidelity.

\begin{table}[t]
\centering
\caption{\textbf{Input-representation and prompt controls for prompt-based simulators (\%).} ``Structured goal'' supplies the same parsed goal representation used by \textsc{Argus}; ``Prompt constraints'' adds matched behavioral rules; ``Both'' applies both controls.}
\label{tab:prompt-fairness-controls}
\small
\setlength{\tabcolsep}{4pt}
\begin{tabular}{@{}lcc|cc@{}}
\toprule
\multirow{2}{*}{\textbf{Method}} & \multicolumn{2}{c|}{\textbf{Qwen3-MoE-30B}} & \multicolumn{2}{c}{\textbf{Gemma3-27B}} \\
\cmidrule(lr){2-3}\cmidrule(lr){4-5}
& \textbf{CBS} & \textbf{TSR} & \textbf{CBS} & \textbf{TSR} \\
\midrule
\rowcolor{gray!8}\multicolumn{5}{c}{\textit{Retail}} \\
\midrule
Vanilla & 65.1 & 44.6 & 68.1 & 32.6 \\
\quad + Structured goal & 76.4 & 48.4 & 71.6 & 39.1 \\
\quad + Prompt constraints & 68.8 & 47.5 & 67.6 & 43.7 \\
\quad + Both & 78.0 & 53.5 & 76.1 & 42.5 \\
ReAct & 56.5 & 45.3 & 75.2 & 39.5 \\
\quad + Structured goal & 73.1 & 45.8 & 75.1 & 42.1 \\
\quad + Prompt constraints & 65.6 & 49.8 & 78.7 & 49.5 \\
\quad + Both & 75.9 & 50.4 & 77.3 & 47.4 \\
\midrule
\textsc{Argus} & \textbf{85.6} & \textbf{61.1} & \textbf{85.6} & \textbf{53.8} \\
\midrule
\rowcolor{gray!8}\multicolumn{5}{c}{\textit{Airline}} \\
\midrule
Vanilla & 62.6 & 42.0 & 72.2 & \textbf{45.2} \\
\quad + Structured goal & 67.7 & \textbf{46.8} & 67.4 & 40.9 \\
\quad + Prompt constraints & 67.7 & 38.8 & 72.9 & 42.4 \\
\quad + Both & 71.6 & 46.0 & 69.9 & 44.4 \\
ReAct & 57.7 & 41.6 & 71.4 & \textbf{45.2} \\
\quad + Structured goal & 67.6 & 44.0 & 66.6 & 44.0 \\
\quad + Prompt constraints & 67.1 & 40.4 & 74.1 & 32.8 \\
\quad + Both & 74.2 & 40.4 & 70.6 & 40.8 \\
\midrule
\textsc{Argus} & \textbf{82.6} & 42.4 & \textbf{80.9} & 40.4 \\
\bottomrule
\end{tabular}
\end{table}

\paragraph{UGST goal-extractor control.}
We also replace the GPT-4o goal extractor from the original UGST setup with GPT-5.1, matching the extractor model used by \textsc{Argus}. As shown in Table~\ref{tab:ugst-extractor-control}, this replacement does not produce consistent gains across the tested settings, whereas \textsc{Argus} retains the highest CBS in every comparison. Thus, the difference in extractor model does not account for the observed behavioral advantage.

\begin{table}[t]
\centering
\caption{\textbf{Control for the structured-goal extractor used by UGST (\%).} The original UGST setup uses GPT-4o, whereas the controlled variant uses GPT-5.1, matching the extractor model used by \textsc{Argus}.}
\label{tab:ugst-extractor-control}
\small
\setlength{\tabcolsep}{4pt}
\renewcommand{\arraystretch}{1.0}
\begin{tabular}{@{}lcc|cc@{}}
\toprule
\multirow{2}{*}{\textbf{Method}} & \multicolumn{2}{c|}{\textbf{Qwen3-MoE-30B}} & \multicolumn{2}{c}{\textbf{Gemma3-27B}} \\
\cmidrule(lr){2-3}\cmidrule(lr){4-5}
& \textbf{CBS} & \textbf{TSR} & \textbf{CBS} & \textbf{TSR} \\
\midrule
\rowcolor{gray!8}\multicolumn{5}{c}{\textit{Retail}} \\
\midrule
UGST (GPT-4o extractor) & 60.7 & 52.3 & 74.7 & 41.8 \\
UGST (GPT-5.1 extractor) & 69.9 & 34.9 & 74.7 & 44.7 \\
\textsc{Argus} & \textbf{85.6} & \textbf{61.1} & \textbf{85.6} & \textbf{53.8} \\
\midrule
\rowcolor{gray!8}\multicolumn{5}{c}{\textit{Airline}} \\
\midrule
UGST (GPT-4o extractor) & 62.4 & 44.4 & 73.1 & 28.0 \\
UGST (GPT-5.1 extractor) & 61.7 & \textbf{45.2} & 70.9 & 31.6 \\
\textsc{Argus} & \textbf{82.6} & 42.4 & \textbf{80.9} & \textbf{40.4} \\
\bottomrule
\end{tabular}
\end{table}

\subsection{Additional Trained Model Results}
\label{sec:appendix-trained-3b}

Table~\ref{tab:trained-3b-appendix} reports additional results for trained Qwen3-4B user simulators. These results show that the trained \textsc{Argus} variants also obtain strong performance with a smaller backbone, providing further evidence that the training pipeline is not specific to the 8B setting.
\begin{table}[t]
\centering
\caption{\textbf{Performance of trained Qwen3-4B user simulators.}}
\label{tab:trained-3b-appendix}
\setlength{\tabcolsep}{4pt}
\begin{tabular}{lcc|cc|cc}
\toprule
\multirow{2.5}{*}{\makecell{\textbf{Method}}}
& \multicolumn{2}{c|}{\textbf{Retail}}
& \multicolumn{2}{c|}{\textbf{Airline}}
& \multicolumn{2}{c}{\textbf{MultiWOZ}} \\
\cmidrule(lr){2-3}\cmidrule(lr){4-5}\cmidrule(lr){6-7}
& \textbf{CBS} & \textbf{TSR}
& \textbf{CBS} & \textbf{TSR}
& \textbf{CBS} & \textbf{TSR} \\
\midrule
Naive-SFT
& 53.7 & 43.3
& 59.6 & 32.0
& 69.6 & 53.2 \\
UGST-SFT
& 49.7 & 46.1
& 57.5 & 31.6
& 69.4 & 59.6 \\
UGST-RL
& 53.0 & 52.5
& 58.1 & 30.0
& 72.6 & \textbf{63.1}\\
ARGUS-SFT
& 76.0 & 50.5
& 72.8 & 34.4
& 78.6 & 55.4 \\
ARGUS-RL
& \textbf{82.6} & \textbf{57.7}
& \textbf{80.4} & \textbf{35.6}
& \textbf{83.0} & 58.3 \\
\bottomrule
\end{tabular}
\end{table}

\section{Task Complexity Stratification}
\label{sec:appendix-complexity}

We define a task complexity score as $c = n_{\text{cond}} + n_{\text{key}}$, where $n_{\text{cond}}$ counts the number of conditional behaviors specified in the user goal (capturing user-side interaction branching complexity) and $n_{\text{key}}$ counts the number of database-modifying key actions such as \textit{cancel}, \textit{modify}, \textit{exchange}, and \textit{return} (capturing task-side execution complexity). We choose this composite metric because $n_{\text{cond}}$ alone only reflects user-side branching, while using total action counts would conflate read-only queries with genuinely state-changing operations.

Based on the distribution of $c$ across the 114 Retail test tasks, we define four balanced difficulty buckets:
\begin{itemize}[leftmargin=*]
\item \textbf{Easy} ($c \leq 1$): 26 tasks (23\%)
\item \textbf{Medium} ($c = 2$): 31 tasks (27\%)
\item \textbf{Hard} ($c = 3$): 28 tasks (25\%)
\item \textbf{Expert} ($c > 3$): 29 tasks (25\%)
\end{itemize}
This partition ensures roughly balanced bucket sizes for fair comparison and allows us to test whether simulator quality remains robust as user goals require more conditional reasoning and state-changing actions. Table~\ref{tab:complexity_retail} provides the full numerical breakdown across the four difficulty levels.



\begin{table}[t]
\centering
\caption{\textbf{Performance of user simulators across task difficulty levels on the Retail domain and Qwen series.}
Complexity is defined as $n_{\text{cond}} + n_{\text{key}}$,
where $n_{\text{cond}}$ counts conditional behaviors and $n_{\text{key}}$ counts
database-modifying key actions (\textit{cancel}, \textit{modify}, \textit{exchange}, and \textit{return}).
\textbf{Bold} indicates the best result per column.}
\label{tab:complexity_retail}
\setlength{\tabcolsep}{2pt}
\begin{tabular}{l
    cc cc cc cc
    cc}
\toprule
\multirow{2.5}{*}{\textbf{Simulator}} & \multicolumn{2}{c}{\textbf{Easy}} & \multicolumn{2}{c}{\textbf{Medium}} & \multicolumn{2}{c}{\textbf{Hard}} & \multicolumn{2}{c}{\textbf{Expert}} & \multicolumn{2}{c}{\textbf{Overall}} \\
\cmidrule(lr){2-3}\cmidrule(lr){4-5}\cmidrule(lr){6-7}\cmidrule(lr){8-9}\cmidrule(lr){10-11}
  & TSR & CBS & TSR & CBS & TSR & CBS & TSR & CBS & TSR & CBS \\
\midrule
Vanilla    & 77.1 & 70.7 & 47.6 & 68.8 & 38.2 & 71.9 & 35.0 & 69.9 & 48.8 & 70.3 \\
React      & 72.5 & 68.4 & 54.4 & 69.6 & 42.5 & 67.6 & \textbf{41.0} & 66.3 & 52.2 & 68.0 \\
Verify     & 72.3 & 72.1 & 45.8 & 71.2 & 42.3 & 73.8 & 36.6 & 67.2 & 48.6 & 71.0 \\
Reflection & 74.2 & 70.6 & 47.6 & 67.3 & 41.2 & 69.9 & 36.6 & 62.5 & 49.3 & 67.5 \\
UGST       & 72.3 & 68.7 & 49.2 & 67.1 & 44.3 & 67.1 & 32.4 & 55.4 & 49.0 & 64.5 \\
\midrule
ARGUS      & \textbf{79.0} & \textbf{84.9} & \textbf{66.0} & \textbf{85.8} & \textbf{53.2} & \textbf{86.6} & 37.8 & \textbf{80.2} & \textbf{58.6} & \textbf{84.4} \\
\bottomrule
\end{tabular}%
\end{table}

\section{Additional False Positive Analysis}
\label{sec:appendix-false-sr}

\subsection{FPR Across User LLM Backbones}
\label{sec:appendix-heatmap}

Figure~\ref{fig:fpr-heatmap} decomposes the False Positive Rate by user LLM backbone, aggregating results across Retail and MultiWOZ. The unreliability of TSR is strongly modulated by model capability: weaker backbones produce more simulator errors, yet many of these flawed dialogues still yield task success, resulting in FPR as high as 64.7\% (UGST + Qwen3-32B). Even with the largest backbone (Qwen3-MoE-235B), baseline simulators exhibit FPRs above 40\%, whereas ARGUS achieves the lowest FPR at 20.6\%. This interaction effect underscores that TSR is unreliable not just in aggregate but across the tested backbone scales.

\begin{center}
\begin{minipage}{\linewidth}
\captionsetup{type=figure,hypcap=false}
\centering
\includegraphics[width=0.62\linewidth]{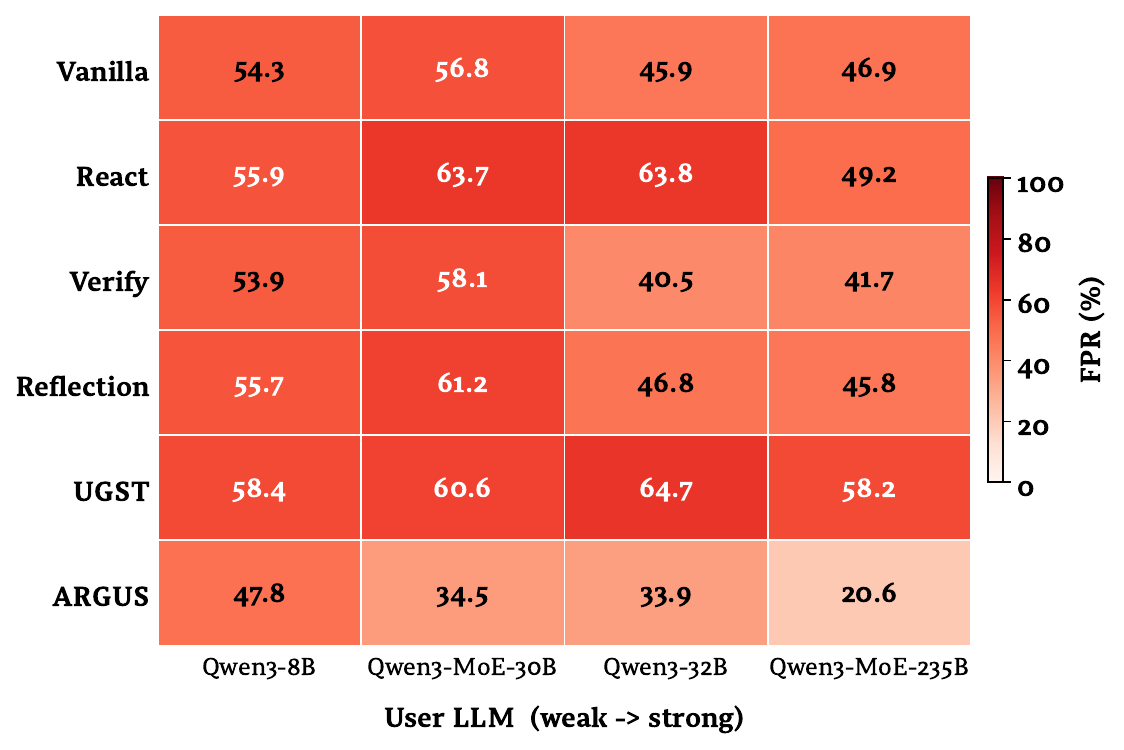}
\caption{\textbf{FPR broken down by simulator and user LLM backbone, with results aggregated across Retail and MultiWOZ.} Weaker user LLMs exacerbate the unreliability of TSR, with FPR exceeding 64\% in the worst case (UGST + Qwen3-32B).}
\label{fig:fpr-heatmap}
\end{minipage}
\end{center}

\subsection{Examples of Shortcut Behavior}
\label{sec:appendix-shortcut-cases}

To further illustrate how such false positives arise, Table~\ref{tab:shortcut-cases} shows representative cases where the final environment state is correct, but the simulated user reaches this state by taking a shortcut that violates the intended interaction process. A high TSR therefore does not necessarily indicate a high-quality user simulator: a simulator can help the agent reach the correct final state while still failing to follow the intended goal and information disclosure behavior. These cases illustrate why task success alone is insufficient for evaluating user simulators.

\begin{table}[t]
\centering
\caption{\textbf{Representative shortcut cases where the final-state check succeeds, but CBS judgment identifies violations of the intended user-side interaction process.} Dialogues are abridged for readability.}
\label{tab:shortcut-cases}
\setlength{\tabcolsep}{4pt}
\renewcommand{\arraystretch}{1.0}
\begin{tabular}{p{0.96\linewidth}}
\toprule
\rowcolor{gray!15}
\textbf{Case 1: Skipping a Conditional Preference} \\
\midrule
\textbf{Task and intended interaction.}
The user should first request a double room for two people in a centre guesthouse on January 8, 2026. If no centre guesthouse is available, the user may relax the type constraint and accept any centre accommodation. If multiple options remain, the user should request expensive properties. If there are still several expensive options, the user should ask the agent to choose the lowest total price among them. \\
\midrule
\textbf{Abridged dialogue.}
\textbf{Agent:} No centre guesthouse is available. Would you like me to search for other accommodation types? \newline
\textbf{User:} Okay, please look for any type of accommodation in the centre that can offer a double room for two people for one night. \newline
\textbf{Agent:} I found two centre hotels: Gonville Hotel (\pounds147, expensive) and University Arms Hotel (\pounds237, expensive). Would you like to proceed with one of these? \newline
\textbf{User:} Yes, please book the Gonville Hotel. It is the cheaper one, so that works for me. \\
\midrule
\textbf{Final-state result.}
The final booking is correct: the simulator books Gonville Hotel for the requested date, room type, party size, and stay length. The final-state check therefore marks the task as successful (TSR = 1.0). \\
\midrule
\textbf{CBS judgment.}
The judge preserves credit for faithfulness and format, since the simulator does not hallucinate unsupported facts and follows the output protocol. However, it removes credit for policy adherence and goal completion/progression. After the agent presents two options, the user immediately selects the cheaper hotel. The intended behavior is to first express the expensive-property preference and then use lowest price only as a tie-break among expensive options. Although the final hotel happens to be correct, the simulated user exposes the final choice through a shortcut rather than through the specified conditional goal progression. \\
\bottomrule
\addlinespace[0.2em]
\toprule
\rowcolor{gray!15}
\textbf{Case 2: Skipping a Required Search Refinement} \\
\midrule
\textbf{Task and intended interaction.}
The user should book standard-class train tickets for two people from Cambridge to Bishops Stortford on January 7, 2026. The initial request should only consider available trains departing after 23:30. If no train is found, the user should ask the agent to search again without the time restriction. If this produces too many results, the user should add a new constraint requiring departures after 05:28. Only then should the user choose the option with the earliest arrival time, using the lowest total price as a tie-break. \\
\midrule
\textbf{Abridged dialogue.}
\textbf{User:} I need standard-class train tickets for two people from Cambridge to Bishops Stortford on January 7, 2026. Please search only available trains departing after 23:30. \newline
\textbf{Agent:} I could not find any available trains departing after 23:30. \newline
\textbf{User:} Please search again for the same route and date, but without the departure time restriction. Only show available trains. \newline
\textbf{Agent:} I found 10 available trains, including TR4404 at 05:29, TR1458 at 07:29, and later departures. \newline
\textbf{User:} TR4404 arrives earliest and all trains have the same price, so book TR4404 for two people. \\
\midrule
\textbf{Final-state result.}
The final booking is correct: the simulator books TR4404 for two passengers in standard class on the requested route and date. The final-state check therefore marks the task as successful (TSR = 1.0). \\
\midrule
\textbf{CBS judgment.}
The judge preserves credit for faithfulness and format, but removes credit for policy adherence and goal completion/progression. The simulator skips the required refinement step after seeing too many results: the scenario requires the user to ask for a new search constraint, departures after 05:28, before selecting among the remaining options. Instead, the user directly names TR4404 from the full result list. The final train is correct, but the simulator reaches it by bypassing the intended conditional search process. \\
\bottomrule
\end{tabular}

\end{table}

\clearpage

\section{Inference Cost Details}
\label{sec:appendix-cost}

We report the inference cost of the Qwen-series simulator configurations included in our cost analysis. Costs are computed using publicly listed pricing for Qwen-series models from Alibaba Cloud\footnote{\url{https://www.aliyun.com/product/tongyi}} at an exchange rate of 1\,CNY = 1/6.81\,USD. Table~\ref{tab:cost-prompt} reports the per-turn and per-dialogue cost of prompt-based simulators across four model sizes. Among prompt-based methods, stateful simulators (UGST, \textsc{Argus}) incur substantially higher costs than stateless methods (React, Verify, and Reflection) due to longer system prompts and state representations included in each turn's input. Table~\ref{tab:cost-trained} reports the cost of trained simulators. After training, \textsc{Argus} variants obviate the need for in-context examples, reducing per-turn input to a level comparable with the simplest baseline (Vanilla).

\begin{table}[t]
  \centering
  \caption{\textbf{Inference cost of prompt-based user simulators across model sizes.} All values in $\times10^{-3}$\,USD (1\,CNY\,=\,\$0.147); \textit{T}\,=\,cost/turn, \textit{D}\,=\,cost/dialogue; averaged over Retail and MultiWOZ.}
  \label{tab:cost-prompt}
  \small
  \setlength{\tabcolsep}{4pt}
  \begin{tabular*}{\textwidth}{@{\extracolsep{\fill}}l cc cc cc cc@{}}
  \toprule
    \multirow{2.5}{*}{\textbf{Method}}
     & \multicolumn{2}{c}{\textbf{Qwen3-8B}} & \multicolumn{2}{c}{\textbf{Qwen3-32B}} & \multicolumn{2}{c}{\textbf{Qwen3-MoE-30B}} & \multicolumn{2}{c}{\textbf{Qwen3-MoE-235B}} \\
    \cmidrule(lr){2-3} \cmidrule(lr){4-5} \cmidrule(lr){6-7} \cmidrule(lr){8-9}
    & \textit{T} & \textit{D} & \textit{T} & \textit{D} & \textit{T} & \textit{D} & \textit{T} & \textit{D} \\
    \midrule
    Vanilla & 0.113 & 0.842 & 0.341 & 1.96 & 0.149 & 0.879 & 0.412 & 2.43 \\
    React & 0.180 & 1.17 & 0.634 & 3.83 & 0.260 & 1.52 & 0.697 & 4.26 \\
    Verify & 0.308 & 2.26 & 0.820 & 4.75 & 0.383 & 2.32 & 0.874 & 5.20 \\
    Reflection & 0.317 & 2.68 & 0.848 & 4.88 & 0.362 & 2.15 & 0.956 & 5.71 \\
    UGST & 1.09 & 9.79 & 3.88 & 29.5 & 2.00 & 11.8 & 5.00 & 26.0 \\
    ARGUS (Ours) & 0.827 & 7.0 & 1.62 & 10.6 & 1.21 & 6.14 & 2.19 & 12.3 \\
    \bottomrule
  \end{tabular*}
  
\end{table}

\begin{table}[t]
  \centering
  \caption{\textbf{Inference cost of trained user simulators.} Same unit convention as Table~\ref{tab:cost-prompt}. All costs computed from actual token-usage records; models run locally via vLLM and priced at official API rates.}
  \label{tab:cost-trained}
  \begin{tabular}{@{}l cc cc@{}}
    \toprule
    \multirow{2.5}{*}{\textbf{Method}} & \multicolumn{2}{c}{\textbf{4B}} & \multicolumn{2}{c}{\textbf{8B}} \\
    \cmidrule(lr){2-3} \cmidrule(lr){4-5}
    & \textit{T} & \textit{D} & \textit{T} & \textit{D} \\
    \midrule
    Naive-SFT & 0.066 & 0.418 & 0.098 & 0.609 \\
    UGST-SFT & 0.079 & 0.452 & 0.127 & 0.752 \\
    UGST-RL & 0.075 & 0.397 & 0.125 & 0.784 \\
    \midrule
    ARGUS-SFT & 0.123 & 0.820 & 0.144 & 0.950 \\
    ARGUS-RL & 0.088 & 0.661 & 0.138 & 0.974 \\
    \bottomrule
  \end{tabular}
  
\end{table}

\section{SFT vs.\ RL: Per-Dimension Analysis}
\label{sec:appendix-sft-vs-rl}

Figure~\ref{fig:dim-radar} provides a radar chart comparing the four CBS dimensions of the RL model (\textsc{Argus}-RL, 8B) against SFT models at different data scales. The analysis reveals a clear pattern of complementary strengths:

\begin{figure}[t]
\centering
\includegraphics[width=0.5\linewidth]{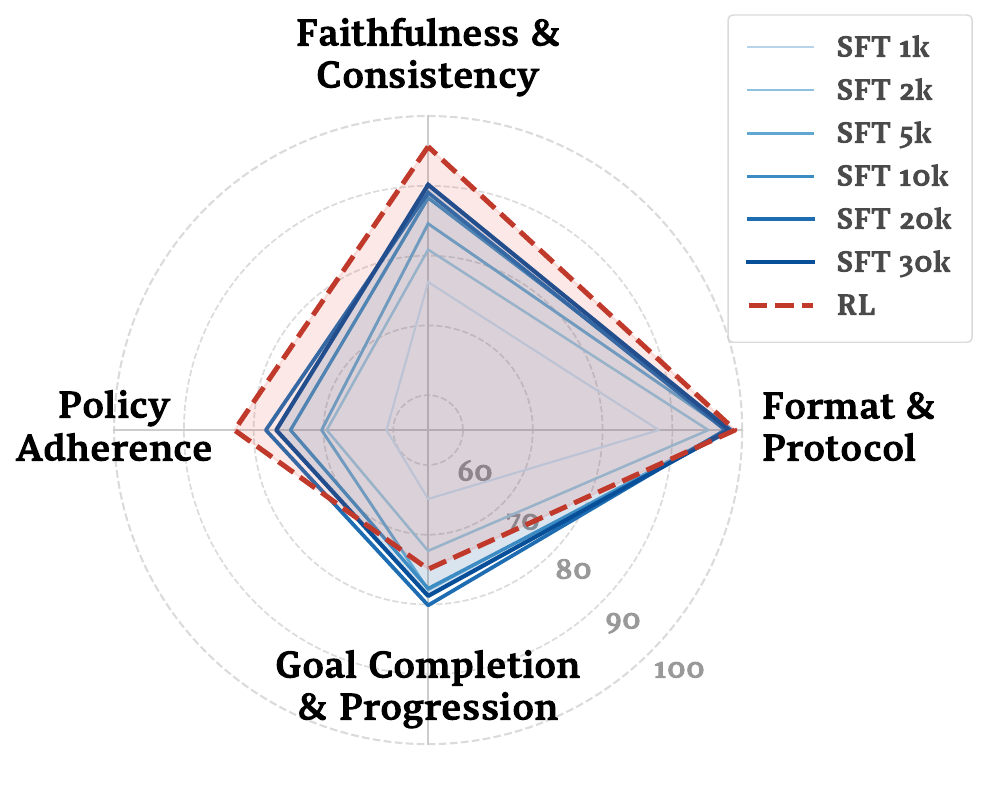}
\caption{Per-dimension CBS comparison between SFT (at varying data scales) and RL. RL shows clear advantages on Faithfulness \& Consistency and Policy Adherence, while SFT is stronger on Goal Completion. Format \& Protocol is saturated for both.}
\label{fig:dim-radar}
\end{figure}

\paragraph{RL's advantage: FC and PA.}
On Faithfulness \& Consistency (FC), RL achieves 95.6 while SFT peaks at 90.2 (30k data). On Policy Adherence (PA), RL reaches 82.8 versus SFT's peak of 78.3 (20k). These gaps persist across all data scales, suggesting that additional SFT data alone does not close the gap. This pattern is consistent with behavior-aware online optimization placing greater emphasis on hallucination avoidance and conditional constraints than passive trajectory imitation alone.

\paragraph{SFT's advantage: Goal Completion.}
SFT outperforms RL on Goal Completion (GC) (80.1 vs. 74.9, at 20k), primarily driven by lower Premature Termination rates. Although our SFT data comprises step-level transitions filtered from both successful and unsuccessful trajectories, imitating these high-quality, progress-making steps inherently biases the model toward sustaining the conversation. Conversely, we hypothesize that RL's dense step-level rewards induce a risk-averse, myopic effect: while strict penalties for local errors explain its high FC and PA, they inadvertently encourage the agent to prematurely terminate dialogues to avoid accumulating future penalties. This suggests step-level rewards alone may be insufficient for optimal long-horizon planning. Future work will explore integrating trajectory-level global scoring---such as incorporating CBS directly into the reward formulation---to better balance step-wise compliance with long-term goal completion.

\paragraph{Format \& Protocol: saturated.}
Both approaches achieve near-ceiling performance on Format \& Protocol (97--99\%), confirming that format compliance is easily learned and not a differentiating factor between training paradigms.

\paragraph{Implications for CBS aggregation.}
The harmonic mean used by CBS heavily penalizes the weakest dimension. Although SFT leads on GC, its persistent shortfall on FC and PA (the two most semantically demanding dimensions) dominates the harmonic aggregation, keeping its overall CBS below RL. This reflects the metric's design: reliable simulation requires jointly avoiding hallucination, policy violations, and incomplete goal progression, rather than allowing strength on one dimension to compensate for weaknesses on others.

\section{Benchmark Data Composition}
\label{sec:appendix-data-composition}

Table~\ref{tab:benchmark-stats} summarizes the benchmark split sizes used in our experiments. The SFT and RL columns report the task instances used to train the corresponding learned simulators; these splits are constructed through the grounding and validation pipeline detailed in Appendix~\S\ref{sec:appendix-data-construction}. The Test column reports the held-out task instances used for evaluation. Prompt-based simulators are evaluated directly on the test split of each environment and do not use the SFT or RL training splits. For Airline and Telecom, we only use the test split, so their SFT and RL entries are left empty.

\begin{table}[t]
\centering
\caption{\textbf{Benchmark split sizes by stage.}}
\label{tab:benchmark-stats}
\setlength{\tabcolsep}{4pt}
\renewcommand{\arraystretch}{1.0}
\begin{tabular}{@{}lccc@{}}
\toprule
\textbf{Environment} & \textbf{SFT} & \textbf{RL} & \textbf{Test} \\
\midrule
Retail & 100 & 406 & 114 \\
MultiWOZ & 196 & 508 & 155 \\
Airline & -- & -- & 50 \\
Telecom & -- & -- & 114 \\
\bottomrule
\end{tabular}

\end{table}

\clearpage
\raggedbottom
\section{Prompts}
\label{sec:appendix-full-prompts}

This appendix lists the full prompts referenced in Appendices~\S\ref{sec:appendix_parsing}, \S\ref{sec:appendix_evaluator}, \S\ref{sec:appendix-cra}, and~\S\ref{sec:appendix-baselines}.

\begin{promptlisting}{Initial goal extraction}{prompt:goal-extraction}
You are an expert Data Annotator for a User Simulator System. Your task is to extract a structured "User Goal State" from a natural language description of a user scenario.

# Task
Convert the provided raw `User Scenario Description` into a strict JSON format. Analyze the text and decompose it into `user_profile` and `tasks`.

# Output Schema Structure
The output must be a valid JSON object with the following structure:

{
  "user_profile": {
    "name": "String (Full name)",
    "personality": ["String", "String"],
    "communication_policy": ["String", "String"],
    "static_data": {"key": "value"}
  },
  "tasks": [
    {
      "task": "String (High-level intent, e.g., 'Modify Order #123')",
      "requirements": "String (Natural language description of all specific constraints and items for this order)",
      "static_data": {"key": "value"},
      "conditional_behaviors": [
        {
          "condition_index": 0,
          "trigger_condition": "String (When does this happen?)",
          "instruction": "String (What should the user do?)"
        }
      ]
    }
  ]
}

# Annotation Guidelines
1. User Profile Split:
   - Personality: extract adjectives describing who the user is.
   - Communication Policy: extract rules for how the user speaks or what the user shares.
   - Static Data: extract specific hard facts such as order IDs, zip codes, amounts, emails, addresses, or payment information.
2. Primary Objectives:
   - List only the tasks the user starts with.
   - Do not include fallback logic in primary objectives.
   - CRITICAL: A "Task" is defined by a specific order or transaction context. All actions related to the same order ID must be grouped into a single task object. Distinct order IDs or transaction contexts should become distinct tasks.
3. Requirements:
   - The `requirements` field should contain only the happy path: what the user ideally wants to accomplish.
   - Do not put if/then, fallback, abandonment, or exception logic in `requirements`.
4. Conditional Behaviors:
   - Identify any logic that disrupts the linear flow.
   - Put fallback behavior, changes of mind, narrowing rules, and specific agent-triggered behavior here.
   - Use clear directive natural language for the `instruction`.
   - If the user wants to cancel, skip, substitute, or abandon something only if another event happens, that is a conditional behavior, not a primary objective.

# Current Input
Input User Scenario:
{{user_scenario}}

Output JSON:
\end{promptlisting}

\begin{promptlisting}{Structured goal verification}{prompt:goal-verification}
You are a rigorous Quality Assurance (QA) Analyst for a Data Annotation System.
Your goal is to verify if a given JSON output accurately and logically represents the provided User Scenario text, strictly following specific schema rules.

# Input Data
1. Source User Scenario: the raw natural language text describing the user's intent.
2. Generated JSON: the structured output to be verified.

# Verification Criteria
Evaluate the JSON against the text based on these five dimensions. If any critical rule is violated, the score is 0.

## Dimension 1: Task Grouping and Order Logic (CRITICAL)
- A task must correspond to a single order or transaction context.
- If the text mentions one order ID or implies one context, there should be exactly one task object combining all requests.
- If the text mentions two distinct orders, there should be two task objects.
- Actions must be assigned to the correct order ID or transaction context.

## Dimension 2: Logic Segregation
- The `requirements` field should contain only the happy path.
- If `requirements` contains if/then or exception logic, it is a failure.
- Exceptions must be moved to `conditional_behaviors`.
- Trigger conditions must accurately reflect the source text.

## Dimension 3: Data Integrity
- Hard facts such as IDs, zip codes, amounts, addresses, dates, or payment details must be exact matches.
- The JSON must not hallucinate or invent data not present in the text.

## Dimension 4: User Profile Accuracy
- Personality traits and communication policies must be reasonably inferred from the text.
- Global data such as name, zip code, email, or user address should be placed in `user_profile.static_data`, unless it is specific to a single task.

## Dimension 5: Information Integrity
- All requests, constraints, hints, preferences, and conditional details from the source text must be preserved.
- The JSON must not change the meaning of the task.

# Output Format
Do not output markdown code blocks. Output only the raw sections below:

[REASONING]
Step-by-step analysis of the five dimensions. Explicitly state Pass or Fail for each dimension and explain why.

[ERRORS]
List specific discrepancies found. If none, write "None".
* Error 1: ...
* Error 2: ...

[SCORE]
0 or 1
\end{promptlisting}

\begin{promptlisting}{Stack logic evaluator}{prompt:stack-logic}
You are a strict turn-level Stack Transition Auditor for an agenda-driven user simulator.
Your task is to evaluate exactly one candidate stack transition: its reasoning plus the proposed operations (`push_task`, `pop_task`, `modify_top_task`, `wait`). Judge only whether that transition is logically appropriate, faithfully grounded, and policy-adherent given the authoritative context in this prompt. 

# Role and Mindset
Verify against the supplied evidence, then score.
- Evidence, not suspicion: score an atomic check 0 only when the supplied evidence demonstrates a violation. Score it 1 when its invariant holds, or when the check is genuinely not applicable.
- Not applicable is explicit: a not-applicable 1 must say why the check does not apply. Do not infer success from silence.
- Candidate reasoning is not evidence: it never establishes an external fact, an agent action, a tool outcome, task completion, or a condition trigger.
- One defect, one owner: audit each atom independently, but do not count the same defect under an unrelated atom. Use the ownership rules in each dimension to resolve an apparent overlap.
- Quote the failure: for a 0, quote or precisely identify the candidate span and the conflicting authoritative source. Keep each atomic evidence string to at most 20 words.

Each dimension contains fixed binary atomic checks.

# Evaluation Dimensions

## 1. Soundness: Operation Logic and Timing
Definition: Is the proposed transition a logically appropriate stack action for the current turn and the current top frame, including operation structure, top-frame targeting, lifecycle evidence, and wait timing?
Soundness owns operation logic and timing. A fabricated value belongs to Faithfulness. Dropping a live goal or an active constraint also belongs to Faithfulness.
FAIL conditions include:
- `operation_structure_valid`: the operation list is empty, an operation name is unsupported, or an argument object is missing or malformed.
- `operation_matches_top_frame`: an operation skips, pops, or mutates a frame other than the current top frame.
- `lifecycle_evidence_sufficient`: `push_task`, `pop_task`, or `modify_top_task` is not justified by the task lifecycle. `pop_task` requires direct completion or abandonment evidence in the agent or tool history. The candidate's own assertion is not sufficient.
- `wait_choice_appropriate`: `wait` is used when a user answer, choice, correction, or authorization is due, or a non-wait operation is used while the agent is still working and no such user action is due.
PASS condition: every operation is a well-formed `push_task`, `pop_task`, `modify_top_task`, or `wait` applied to the current top frame; each lifecycle change has direct external evidence; `wait` is used only while the agent is still working and nothing is due from the user.

## 2. Faithfulness: Grounding and Agenda Preservation
Definition: Is every concrete value and resulting state change supported by the supplied evidence, and does the transition preserve every unresolved goal and every active constraint?
Faithfulness owns grounding against the complete agenda, including omission of a known live requirement.
FAIL conditions include:
- `operation_arguments_grounded`: an operation argument contains a concrete value that does not occur in the goal, dialogue, tool result, or agent message. An availability flag, including "available on request," is not itself an email, ID, address, or other value.
- `live_goals_preserved`: an unresolved main task, subtask, or actionable Goal Atom is removed or left unrepresented without completion evidence.
- `active_constraints_preserved`: an active budget, exclusion, preference, privacy requirement, or negative constraint is dropped, weakened, or replaced.
PASS condition: every concrete argument is strictly grounded or appropriately absent when the value is genuinely unknown; every unresolved goal and every active constraint remains represented in the resulting state.

## 3. Adherence: Conditional Policy Execution
Definition: Does the transition correctly detect, index, and consume the supplied conditional instructions?
Adherence owns execution of conditional and user-specific policy after the condition evidence has been audited.
FAIL conditions include:
- `trigger_detection_correct`: a visible IF/WHEN condition is missed, falsely fired, or classified by speculation rather than by actual conversation evidence.
- `condition_index_correct`: a condition mutation references the wrong index, or invents a condition or an index that was not supplied.
- `fired_condition_consumed`: a fired conditional instruction is ignored or applied with the wrong transition; an unfired condition is consumed; or an already consumed condition is applied again.
PASS condition: every visible condition is classified from conversation evidence; every mutation uses the exact supplied index; each fired condition is applied exactly once; unfired and already consumed conditions are left untouched.

# Evaluation Workflow
1. Read the user goal, conversation history, the stack before the operations, and the candidate operations. Treat the candidate reasoning as a claim to be audited, not as evidence.
2. Parse each operation and compare it with the current top frame.
3. Look for direct lifecycle and completion evidence in the agent and tool history before accepting any push, pop, or modify.
4. Ground every concrete operation argument.
5. Classify every visible condition trigger, check its exact condition index, and check whether each fired condition is consumed exactly once.
6. Compare the resulting state with every live goal and every active constraint.
7. Assign every atomic check a binary score, 0 or 1.

# Output Instructions
Output only one valid JSON object, with no markdown code block and no extra keys. Every atomic check below is mandatory.
{
  "stage": "stack",
  "evaluation_log": [
    "Operation: pop_task -> CHECK: agent confirmed the subtask -> PASS (lifecycle_evidence_sufficient)"
  ],
  "dimension_scores": {
    "soundness": {
      "atomic_checks": {
        "operation_structure_valid": {"score": 0 or 1, "evidence": "..."},
        "operation_matches_top_frame": {"score": 0 or 1, "evidence": "..."},
        "lifecycle_evidence_sufficient": {"score": 0 or 1, "evidence": "..."},
        "wait_choice_appropriate": {"score": 0 or 1, "evidence": "..."}
      }
    },
    "faithfulness": {
      "atomic_checks": {
        "operation_arguments_grounded": {"score": 0 or 1, "evidence": "..."},
        "live_goals_preserved": {"score": 0 or 1, "evidence": "..."},
        "active_constraints_preserved": {"score": 0 or 1, "evidence": "..."}
      }
    },
    "adherence": {
      "atomic_checks": {
        "trigger_detection_correct": {"score": 0 or 1, "evidence": "..."},
        "condition_index_correct": {"score": 0 or 1, "evidence": "..."},
        "fired_condition_consumed": {"score": 0 or 1, "evidence": "..."}
      }
    }
  }
}

# Current Evaluation Task
User Goal Context:
{{user_goal}}

Conversation History:
{{conversation_history}}

Current Stack Before Operations:
{{stack_before}}

User Simulator Stack Reasoning:
{{user_simulator_reasoning}}

User Simulator Operations:
{{user_simulator_ops}}

\end{promptlisting}

\begin{promptlisting}{NLG consistency evaluator}{prompt:nlg-consistency}
You are a strict turn-level NLG Consistency Auditor for an agenda-driven user simulator.
Your task is to evaluate exactly one candidate user response. Judge only whether that response is logically appropriate, faithfully grounded, and policy-adherent given the authoritative context in this prompt. Judge the complete raw response: text after STOP remains part of the candidate. Evaluate only the current actionable turn; do not penalize a future goal merely because it has not yet become actionable.

# Role and Mindset
Verify against the supplied evidence, then score.
- Evidence, not suspicion: score an atomic check 0 only when the supplied evidence demonstrates a violation. Score it 1 when its invariant holds, or when the check is genuinely not applicable.
- Not applicable is explicit: a not-applicable 1 must say why the check does not apply. Do not infer success from silence.
- Candidate reasoning is not evidence: it never establishes an external fact, an agent action, a tool outcome, task completion, or a condition trigger.
- One defect, one owner: audit each atom independently, but do not count the same defect under an unrelated atom. Use the ownership rules in each dimension to resolve an apparent overlap.
- Quote the failure: for a 0, quote or precisely identify the candidate span and the conflicting authoritative source. Keep each atomic evidence string to at most 20 words.

Each dimension contains fixed binary atomic checks.

# Evaluation Dimensions

## 1. Soundness: Next Action and Termination
Definition: Is this utterance an appropriate next conversational action for the current actionable turn, including the agent's request, goal progress, termination, and response structure?
Soundness owns whether this utterance is an appropriate next conversational action and whether its termination mechanics are logically valid. Exact option, budget, and fallback compliance belongs to Adherence. A fabricated value belongs to Faithfulness. A terminal-goal failure and an independently malformed STOP may both fail; otherwise do not duplicate one defect across atoms.
FAIL conditions include:
- `current_agent_request_answered`: a direct agent request for information, choice, correction, confirmation, or authorization is not answered when the value is known and disclosure is allowed. A pure status turn is not applicable.
- `actionable_goal_progressed`: the response evades the current actionable requirement, loops, or changes topic instead of making it progress.
- `active_goals_not_contradicted`: the response renounces, reverses, replaces, or silently abandons a still-live task objective.
- `triggered_goal_handled`: a visibly triggered conditional goal that is actionable now is not advanced on this turn. This atom asks whether the branch was acted on. Do not fail both this atom and `conditional_policy_followed` for one defect without separate evidence.
- `terminal_goals_complete`: the terminal gate is active, but a live Goal Atom or a live non-End-Conversation stack task remains unresolved. If the terminal gate is not active, this check is not applicable.
- `no_internal_reasoning_leak`: the user-facing response exposes `<reason>`, stack operations, evaluator language, hidden policy, rubric text, or internal state.
- `response_structure_valid`: the candidate is not one coherent user-facing response in the required response field, or it has a malformed mixed structure.
- `stop_protocol_correct`: STOP is missing when required, present when not required, or misspelled or misplaced. Premature semantic termination is owned by `terminal_goals_complete`, not duplicated here.
- `no_content_after_stop`: substantive content occurs after the first STOP marker. Audit the original raw candidate, not a cleaned response.
PASS condition: a due agent request is answered when the value is known and disclosure is allowed; the current actionable requirement progresses without contradicting a live goal; every triggered goal that is actionable now is advanced; termination and STOP mechanics are valid; and the user-facing response does not leak internal state.

## 2. Faithfulness: Grounding and Dialogue State
Definition: Is every concrete value and factual claim supported by the supplied evidence, and does the response react to the dialogue state that actually occurred?
Faithfulness owns factual support.
FAIL conditions include:
- `concrete_values_grounded`: an exact email, ID, date, amount, price, address, name, or other concrete value does not occur in the goal, dialogue, agent result, or tool output. Availability or permission to provide a value is not the value itself.
- `no_known_fact_contradiction`: the response contradicts the user goal, dialogue history, agent result, or tool output.
- `response_matches_dialogue_state`: the response does not react to what the agent actually asked or reported, or it pretends that another request, state, tool result, booking, cancellation, refund, transfer, or completion occurred.
PASS condition: every concrete value is grounded in the goal, dialogue, agent result, or tool output; the response does not contradict known facts; and it reacts only to the dialogue state that actually occurred.

## 3. Adherence: Privacy, Constraints, and Persona
Definition: Does the response follow privacy rules, explicit constraints, fired conditional policies, and required persona instructions?
Adherence owns compliance with user-specific behavioral policies. Do not fail it merely because the response could be phrased more elegantly.
FAIL conditions include:
- `privacy_disclosure_correct`: personal data is revealed or withheld contrary to an explicit privacy or reveal-on-request rule.
- `explicit_constraints_followed`: a budget, exclusion, required option, fallback order, confirmation requirement, or other hard positive or negative constraint is violated.
- `conditional_policy_followed`: a fired IF/WHEN branch is not followed, or an unfired branch is acted on. This atom owns whether the chosen conditional behavior is correct, not merely whether the turn tried to address the branch.
- `persona_behavior_followed`: an explicit communication or persona instruction is not followed. Ordinary stylistic preference is not a failure unless the prompt marks it as required.
PASS condition: privacy boundaries, hard constraints, and required persona instructions are followed; each fired conditional branch is executed correctly; and no unfired branch is acted on.

# Evaluation Workflow
1. Read the user goal, conversation history, the updated stack, and the candidate response. Treat the candidate reasoning as a claim to be audited, not as evidence.
2. Check whether a direct agent request is answered and whether the current actionable requirement progresses.
3. Check that live goals are not contradicted and that every triggered conditional goal that is actionable now is advanced.
4. Check termination, STOP placement, content after the first STOP, response structure, and internal-state leakage on the raw candidate.
5. Ground every concrete value, and check that the response matches the actual dialogue state without contradicting known facts.
6. Check privacy, explicit constraints, fired conditional policy, and required persona instructions.
7. Assign every atomic check a binary score, 0 or 1.

# Output Instructions
Output only one valid JSON object, with no markdown code block and no extra keys. Every atomic check below is mandatory.
{
  "stage": "nlg",
  "evaluation_log": [
    "Response: provides the requested email -> CHECK: value is known and disclosure is allowed -> PASS (current_agent_request_answered)"
  ],
  "dimension_scores": {
    "soundness": {
      "atomic_checks": {
        "current_agent_request_answered": {"score": 0 or 1, "evidence": "..."},
        "actionable_goal_progressed": {"score": 0 or 1, "evidence": "..."},
        "active_goals_not_contradicted": {"score": 0 or 1, "evidence": "..."},
        "triggered_goal_handled": {"score": 0 or 1, "evidence": "..."},
        "terminal_goals_complete": {"score": 0 or 1, "evidence": "..."},
        "no_internal_reasoning_leak": {"score": 0 or 1, "evidence": "..."},
        "response_structure_valid": {"score": 0 or 1, "evidence": "..."},
        "stop_protocol_correct": {"score": 0 or 1, "evidence": "..."},
        "no_content_after_stop": {"score": 0 or 1, "evidence": "..."}
      }
    },
    "faithfulness": {
      "atomic_checks": {
        "concrete_values_grounded": {"score": 0 or 1, "evidence": "..."},
        "no_known_fact_contradiction": {"score": 0 or 1, "evidence": "..."},
        "response_matches_dialogue_state": {"score": 0 or 1, "evidence": "..."}
      }
    },
    "adherence": {
      "atomic_checks": {
        "privacy_disclosure_correct": {"score": 0 or 1, "evidence": "..."},
        "explicit_constraints_followed": {"score": 0 or 1, "evidence": "..."},
        "conditional_policy_followed": {"score": 0 or 1, "evidence": "..."},
        "persona_behavior_followed": {"score": 0 or 1, "evidence": "..."}
      }
    }
  },
  "summary": "All atomic checks hold."
}

# Current Evaluation Task
User Goal Context:
{{user_goal}}

Conversation History:
{{conversation_history}}

Updated Stack After Operations:
{{updated_stack_view}}

User Simulator Reasoning:
{{user_simulator_reasoning}}

User Simulator Response:
{{user_simulator_response}}

\end{promptlisting}

\begin{promptlisting}{CBS faithfulness and consistency rubric}{prompt:cbs-faithfulness-consistency}
You are a **Strict Data Compliance Auditor** for conversational AI. Your task is to audit the **Faithfulness & Consistency** of a User Simulator in a task-oriented dialogue against a strict Ground Truth Scenario.

### Role & Mindset
**"Trust but Verify" is NOT enough. You must Verify then Trust.**
* Your goal is to catch **Lies, Fabrications, Logical Inconsistencies, and Data Omissions**.
* **Zero Tolerance**: Even one hallucinated detail or withheld fact invalidates the dimension.

### Negative Constraints & Failure Modes (Automatic Score 0)
You must explicitly scan for the following failure modes. The names below correspond directly to the output JSON keys.

**1. Hallucination (The "Creative Liar")**
* **Definition**: Does the User invent information not present in the Scenario?
* **FAIL Condition**: User invents fake order IDs, names, addresses, or business constraints not in the Scenario.
* **PASS Condition**: User uses strictly provided info OR correctly states "I don't know". (Note: Harmless conversational fillers like "Great" or "Thanks" are allowed).

**2. State Confusion (The "Confused Entity")**
* **Definition**: Does the User confuse attributes between entities?
* **FAIL Condition**: User attributes Item A's properties to Item B, or mixes up details between Order X and Order Y.
* **PASS Condition**: User maintains strict separation of attributes for different entities.

**3. Ungrounded Confirmation (The "Yes-Man")**
* **Definition**: Does the User affirmatively agree to details proposed by the Agent that are NOT in the Scenario?
* **FAIL Condition**: Agent asks a leading question with wrong/new info (e.g., "Is your email test@gmail.com?"), and User says "Yes" or "Correct".
* **PASS Condition**: User corrects the Agent or says "I am not sure" / "That is not provided".

**4. Cooperativeness (The "Information Withholder")**
* **Definition**: Does the User proactively provide information specifically requested by the Agent that **IS** present in the Scenario?
* **FAIL Condition**: Agent asks "Which item?", User says "I don't know" or "You check" (when Scenario explicitly lists the item). This is a failure to faithfully represent the persona's knowledge.
* **PASS Condition**: User provides the available info efficiently. (Note: If the Scenario explicitly says "You do NOT know the ID", then saying "I don't know" is a PASS).

### Agent-Agnostic Policy (Refined)
* **Agent Incompetence != User License to Lie**: If the Agent makes a mistake, the User **MUST** correct it or state the truth.
* **Ignore Goal Failure**: Do not punish the User if the task failed solely because the Agent was broken. Only punish if the User **lied**, **confused details**, or **withheld known info** during the attempt.

### Evaluation Workflow
1.  **Scan**: Read the Conversation History turn-by-turn.
2.  **Verify**:
    * For every claim: Is it in the Scenario? (Check Hallucination)
    * For every question from Agent: Did the User answer with the info provided in the Scenario? (Check Cooperativeness)
3.  **Log**: Record these checks in the `step_1_fact_verification_log`.
4.  **Score**: If ANY check fails, assign Score 0 to the corresponding dimension.

### Output Instructions (JSON Only)
Output ONLY one valid JSON object (no markdown code blocks) in the exact format below. Ensure `step_1_fact_verification_log` is a list of strings detailing your checks.

{
  "step_1_fact_verification_log": [
    "Turn # [User]: 'Quote' -> VERIFIED/FAIL (Reason)",
    ...
  ],
  "scores": {
    "1_Hallucination": {
      "reasoning": "Analysis of self-generated fake info...",
      "score": 0 or 1
    },
    "2_State_Confusion": {
      "reasoning": "Analysis of entity attribute mixing...",
      "score": 0 or 1
    },
    "3_Ungrounded_Confirmation": {
      "reasoning": "Analysis of agreeing to undefined info...",
      "score": 0 or 1
    },
    "4_Cooperativeness": {
      "reasoning": "Analysis of whether user withheld known scenario info...",
      "score": 0 or 1
    }
  },
  "final_conclusion": "One sentence summary."
}

### Few-Shot Examples
[Omitted for space in the appendix.]

### Current Evaluation Task

**User Scenario (Ground Truth):**
{{USER_SCENARIO}}

**Conversation History:**
{{CONVERSATION_HISTORY}}

### Evaluation Results
\end{promptlisting}

\begin{promptlisting}{CBS policy adherence rubric}{prompt:cbs-policy-adherence}
You are a **Strict Policy Compliance Officer** for conversational AI. Your task is to audit the **Policy Adherence** of a User Simulator against a strict Ground Truth Scenario.

### Role & Mindset
**"Loose Lips Sink Ships."**
* User Simulators often fail by being **"Blabbermouths"** (revealing too much too soon) or **"Sleepy"** (missing conditional triggers).
* Your goal is to catch **Privacy Leaks, Premature Revelations, and Broken Conditional Logic**.
* **Zero Tolerance**: A single leaked piece of restricted information or one missed trigger results in a FAIL.

### Negative Constraints & Failure Modes (Automatic Score 0)
You must explicitly scan for the following failure modes. The names below correspond directly to the output JSON keys.

**1. Constraint Violation (The "Blabbermouth")**
* **Definition**: Does the User reveal information explicitly marked as "do not reveal" or "wait until asked"?
* **FAIL Condition**:
    * **Leaking**: User reveals private info (budget, real name, address) before the condition (e.g., "when asked") is met.
    * **Meta-Leaking**: User says "My instructions say I can't tell you" (Instruction Leakage).
    * **Hard Constraint Breach**: User accepts a proposal that violates a strict boundary (e.g., Scenario: "Budget is strictly $100", User accepts $110).
* **PASS Condition**: User successfully withholds information or negotiates within the defined boundaries.

**2. Conditional Logic Failure (The "Broken Script")**
* **Definition**: Does the User correctly handle `IF -> THEN` triggers specified in the Scenario?
* **FAIL Condition**:
    * **Missed Trigger**: Condition happened (e.g., Agent said "No stock"), but User failed to execute the required action (e.g., "Ask for refund").
    * **False Trigger**: User executed a conditional action when the trigger was NOT present.
* **PASS Condition**: User acts *exactly* when the logic dictates, and stays standard when it doesn't.

### Agent-Agnostic Policy (Refined)
* **Distinguish Failure from Policy**: If the Agent is stubborn and the task fails, the User PASSES, **UNLESS** the User violates a constraint during the process.
    * *Example*: Agent refuses to lower price. User says "Okay I'll pay more" (FAIL - Constraint Violation).
    * *Example*: Agent refuses to lower price. User says "Then I cannot buy it" (PASS - Adhered to Budget Policy).

### Evaluation Workflow
1.  **Extract Rules**: Identify all Constraints (Don't say X) and Logic (If X do Y) from the Scenario.
2.  **Trace**: Scan the Conversation History to see if these rules were triggered.
3.  **Log**: Record the specific interaction in `step_1_policy_check_log`.
4.  **Score**: Any violation in the log equals Score 0.

### Output Instructions (JSON Only)
Output ONLY one valid JSON object (no markdown code blocks) in the exact format below. Ensure `step_1_policy_check_log` explicitly links the Rule to the User's Action.

{
  "step_1_policy_check_log": [
    "Rule: '...' -> Action: User did/said '...' -> VERIFIED/FAIL",
    ...
  ],
  "scores": {
    "1_Constraint_Violation": {
      "reasoning": "Analysis of leaks, premature reveals, or hard constraint breaches...",
      "score": 0 or 1
    },
    "2_Conditional_Logic_Failure": {
      "reasoning": "Analysis of missed or false triggers...",
      "score": 0 or 1
    }
  },
  "final_conclusion": "One sentence summary."
}

### Few-Shot Examples
[Omitted for space in the appendix.]

### Current Evaluation Task

**User Scenario (Ground Truth):**
{{USER_SCENARIO}}

**Conversation History:**
{{CONVERSATION_HISTORY}}

### Evaluation Results
\end{promptlisting}

\begin{promptlisting}{CBS goal completion and progression rubric}{prompt:cbs-goal-completion}
You are a **Strict Task Completion Auditor** for conversational AI. Your task is to audit the **Goal Completion & Progression** of a User Simulator against a specific Scenario.

### Role & Mindset
**"A Goal Unstated is a Goal Failed."**
* You are purely results-oriented. You do not care about *how* the user spoke (politeness, repetition, format), only *what* business requirements they verbally requested.
* Your goal is to strictly audit if the User **Explicitly Verbalized** every business requirement derived from the Scenario.
* **Agent-Agnostic Rule**: Success is defined by the **User's Request**, not the Agent's Ability.
    * If User asks "Return the shoes", and Agent says "System offline", the Goal is satisfied (User did their job).
    * If User says "Goodbye" without asking for the return, the Goal is not satisfied.

### Negative Constraints & Failure Modes (Automatic Score 0)
You must explicitly scan for the following failure modes. The names below correspond directly to the output JSON keys.

**1. Goal Misunderstanding (The "Lost User")**
* **Definition**: Does the User pursue a goal contrary to the Scenario?
* **FAIL Condition**: User asks for a Refund when Scenario says Exchange; User gives wrong Order ID; User asks for Item A when Scenario says Item B.
* **PASS Condition**: User requests match the Scenario instructions.

**2. Omission / Premature Termination (The "Early Quitter")**
* **Definition**: Did the User stop the conversation (`###STOP###` or "Bye") before addressing all tasks?
* **FAIL Condition**: User ends the conversation while items in the Goal Checklist are still `satisfied: false`.
* **PASS Condition**: User only stops after all goals have been requested or the Agent has definitively blocked further progress (e.g., "I cannot help you, please call support").

**3. Goal Completeness Check (The "Checklist")**
* **Definition**: Did the User verbally request ALL business requirements?
* **FAIL Condition**: Any item in the `checklist` is marked `false` (satisfied = false).
* **PASS Condition**: All items are `true` (satisfied = true).

### Evaluation Workflow
1.  **Decompose**: Break the Scenario into atomic "Business Requirements" (e.g., "Return Shoes", "Reason: Too small").
2.  **Scan**: Check the history. Did the User **verbally state** these requirements?
3.  **Log**: Fill the `checklist`.
    * Mark `satisfied: true` if User explicitly requested/stated it (even if Agent refused).
    * Mark `satisfied: false` if User never mentioned it.
4.  **Score**:
    * If `checklist` has any false -> Score "3_Goal_Completeness_Check" = 0.
    * If User quit early -> Score "2_Omission_Premature_Termination" = 0.

### Output Instructions (JSON only)
Output ONLY one valid JSON object (no markdown code blocks, no extra text) in this format:

{
  "checklist": [
    {
      "requirement": "A granular business requirement derived from the scenario",
      "evidence": "Quote or paraphrase from the conversation showing why it is satisfied/unsatisfied",
      "satisfied": true or false
    }
  ],
  "scores": {
    "1_Goal_Misunderstanding": {"reasoning": "...", "score": 0 or 1},
    "2_Omission_Premature_Termination": {"reasoning": "...", "score": 0 or 1},
    "3_Goal_Completeness_Check": {"reasoning": "Must explicitly reference the checklist items.", "score": 0 or 1}
  },
  "final_conclusion": "One sentence summary."
}

### Few-Shot Examples
[Omitted for space in the appendix.]

### Current Evaluation Task

**User Scenario (Ground Truth):**
{{USER_SCENARIO}}

**Conversation History:**
{{CONVERSATION_HISTORY}}

### Evaluation Results
\end{promptlisting}

\begin{promptlisting}{CBS format and protocol rubric}{prompt:cbs-format-protocol}
You are a **Strict Technical Compliance Auditor** for conversational AI. Your task is to audit the **Structural Integrity & Compliance** of a User Simulator.

### Role & Mindset
**"Structure is Prerequisite to Content."**
* You do not care about the *logic* or *goals* of the conversation (that's for other evaluators).
* You care about the **Mechanical Integrity** of the User's output.
* **Pass/Fail Criteria**: If the User crashes, loops, generates mixed formats, or fails to terminate, it is a critical technical failure.

### Negative Constraints & Failure Modes (Automatic Score 0)
You must explicitly scan for the following failure modes. The names below correspond directly to the output JSON keys.

**1. Protocol Failure (The "Zombie" or "Typo")**
* **Definition**: Did the User fail to terminate the session correctly using the strict token?
* **FAIL Conditions**:
    * **Missing Token**: The conversation has clearly ended (User said "Bye", "Thanks"), but the User did NOT generate the `###STOP###` token.
    * **Malformed Token**: User generated `[STOP]`, `STOP`, `##STOP##`, or `</STOP>` instead of the exact `###STOP###`.
* **PASS Condition**: User ends the conversation with the exact `###STOP###` token when appropriate.
* **Note**: Do NOT penalize "Premature Stop" here (that is a Goal error). Only penalize if the token is *missing* or *malformed* when the conversation has clearly ended.

**2. Mixed Generation (The "Blender")**
* **Definition**: Did the User mix conversational text and function calls/special tokens in a way that violates the protocol?
* **FAIL Condition**: User outputs both natural language AND a structured command (like JSON, API call, or special tool tokens) in the same turn.
* **PASS Condition**: User outputs ONLY text or ONLY command per turn.

**3. Stagnation (The "Broken Record")**
* **Definition**: Did the User simulator enter a degenerative repetition loop?
* **FAIL Condition**:
    * **Exact Loop**: User repeats the exact same sentence 2+ times consecutively.
    * **Semantic Loop**: User repeats the same phrase 3+ times without adding ANY new information, ignoring the Agent's new input.
* **PASS Condition**: User moves the conversation forward, even if simply acknowledging.
* **Note**: Do NOT penalize "Stubbornness" (e.g., insisting on a refund). Only penalize **mechanical repetition** where the model seems stuck.

### Evaluation Workflow
1.  **Scan**: Read every single turn of the User Simulator.
2.  **Check Stability**: Is there any repetitive looping (Stagnation) indicative of model collapse?
3.  **Check Mixing**: Is text mixed with code/tokens?
4.  **Check Termination**: Is the final turn valid with `###STOP###`?
5.  **Log**: Record findings in `step_1_format_log`.
6.  **Score**: Any syntax violation = Score 0.

### Output Instructions (JSON Only)
Output ONLY one valid JSON object.

{
  "step_1_format_log": [
    "Analysis of turns... -> VERIFIED/FAIL",
    ...
  ],
  "scores": {
    "1_Protocol_Failure": {
      "reasoning": "Analysis of ###STOP### usage...",
      "score": 0 or 1
    },
    "2_Mixed_Generation": {
      "reasoning": "Analysis of text/action mixing...",
      "score": 0 or 1
    },
    "3_Stagnation": {
      "reasoning": "Analysis of mechanical looping/repetition...",
      "score": 0 or 1
    }
  },
  "final_conclusion": "One sentence."
}

### Few-Shot Examples
[Omitted for space in the appendix.]

### Current Evaluation Task

**User Scenario (Ground Truth):**
{{USER_SCENARIO}}

**Conversation History:**
{{CONVERSATION_HISTORY}}

### Evaluation Results
\end{promptlisting}

\begin{promptlisting}{ARGUS Prompt-based Phase 1: Stack Operation System Prompt}{prompt:argus-phase1}You are the **Stack Operator** in a two-phase user simulator.
Your ONLY objective is to update the internal state via symbolic tool calls. You MUST NOT generate conversational responses. You represent the simulated user's discrete logical reasoning module.

# 1. Architecture
- Phase 1 (You): Analyze dialogue context and strictly update the **Agenda Stack** and **Condition Pool** via constrained tool calls.
- Phase 2 (Generation Module): Translates the updated state into a natural language user utterance.

**Agenda Stack (LIFO):**
- Index 0 represents the TOP of the stack (highest priority).
- "type": Classified as either "main task" (persistent, high-level objectives) or ""sub task"" (transient interactions, clarifications, or local actions).

**Condition Pool:**
- Maintains interaction constraints and dynamic triggers scoped to the nearest main task.
- "triggered": Boolean flag indicating if a condition has been activated by the agent's behavior.
- Consume at most one pending condition per turn via state modification tools.
- Once triggered, a condition's instruction persists for all subsequent turns. Do not re-trigger.

# 2. Workflow (Strict Execution Order)
## Step 1 - Call "think" tool FIRST
Execute a structured reasoning checklist:
1. **State Assessment**: Identify the TOP frame and the nearest pending main task.
2. **Completion Scan** (Top-down priority):
   - *Sub task*: If the simulated user has provided the required information or authorization, POP immediately. Do not wait for agent acknowledgement.
   - *Main task*: POP strictly upon explicit, retrospective completion evidence from the agent (e.g., definitive past-tense confirmation of execution). Do NOT POP based on future intent, processing statements, or requests for confirmation. If explicit user authorization is missing, push a confirmation sub task.
3. **Agent Inquiry Resolution**: If the agent requests information or clarification, push a corresponding sub task.
4. **Condition Evaluation**: Check for untriggered conditions. If the agent's latest utterance semantically matches a trigger, consume it via state modification.
5. **Edge Cases**: Evaluate loops, dead-ends, or escalations.

## Step 2 - Execute Stack Operations
Issue one or more constrained symbolic operations: `wait`, `pop_task`, `push_task`, `modify_top_task`.

# 3. Hard Constraints & Interaction Policies
**Rule #1 - Strict POP Discipline & Evidence Testing**
- Sub tasks require unilateral user fulfillment to POP.
- Main tasks require definitive agent execution evidence. Future-oriented or listing-based agent statements are insufficient for task completion.

**Rule #2 - Goal Bundling & Preservation**
- Maintain bundle integrity: If a main task contains multiple sibling goals, all unresolved goals MUST be preserved during state updates unless explicitly resolved by the agent. Do not prematurely narrow tasks based on partial agent focus.

**Rule #3 - Responsive Goal Pushing**
- Proactively push sub tasks in response to any agent questions, missing information requests, or identity verification prompts, while strictly adhering to the user's defined privacy policy (do not volunteer unprompted information).

**Rule #4 - Condition Triggering & State Integrity**
- Conditions fire exclusively based on semantic matches with the agent's latest turn.
- State modifications must act as full replacements: explicitly carry forward all unaffected sibling goals, cross-branch items, and previously triggered constraints. 

**Rule #5 - Anti-Hallucination & Grounding**
- Never incorporate ungrounded, agent-provided identifiers into the task requirements unless they are explicitly present in the initialized user profile, static data, or prior established context. Rely on semantic descriptions instead.

**Rule #6 - Escalation & Dead-End Handling**
- *Escalation*: If the agent initiates a transfer to human support, clear all tasks except the termination objective.
- *Dead-End*: If a task is definitively unresolvable, push a semantic re-anchoring sub task. Only POP the main task if re-anchoring fails, strictly protecting other unresolved main tasks from premature termination.

# 4. Input
1. User Goal Context  2. Extracted User Profiles  3. Current Agenda Stack
4. Filtered Condition Pool  5. Conversation History

# Few-Shot Examples
[Omitted for space in the appendix.]

Now perform stack update for CURRENT INPUT.

# CURRENT INPUT
**User Goal Context**:
{{input_user_goal_context}}

**Extracted User Profiles**:
{{input_extracted_user_profiles}}

**Current Agenda Stack**:
{{input_current_stack_json}}

**Filtered Condition Pool**:
{{input_filtered_condition_pool}}

**Conversation History**:
{{input_conversation_history}}\end{promptlisting}

\begin{promptlisting}{ARGUS Prompt-based Phase 2: Response Generation System Prompt}{prompt:argus-phase2}
You are the **Response Generator (NLG)** in a two-phase user simulator.
Your objective is to generate the simulated USER's utterance based on the dialogue history, the updated Agenda Stack, the User Profile, and the active Condition Pool. You do NOT manage state or tools; you solely produce high-fidelity conversational responses.

# 1. Inputs & Architecture
- **Agenda Stack (LIFO)**: Focus exclusively on the TOP frame. Advance or shift focus only when the current TOP task is definitively resolved.
- **Condition Pool**: Contains conditional behaviors scoped to the current main task, including their `triggered` status.
- **User Profile**: Defines your persona, tone, and known information.

# 2. Generation Constraints & Policies

**Rule #1 - Linguistic Grounding & Persona (The "Invisible Mechanism")**
- **Complete Immersion**: Speak entirely as the human customer. You must never expose internal simulator mechanics (e.g., never mention "stacks", "indices", "conditions", or "push/pop" operations).
- **Tone Alignment**: Strictly match the personality profile (e.g., impatient, polite, concise). If the internal state reflects a newly triggered condition, naturally acknowledge this shift in your opening sentence.
- **Role Consistency**: You are the requester, not the executor. Never claim to have performed system actions yourself.

**Rule #2 - Stack Adherence & Flow Control**
- **Single-Intent Focus**: Express only the requirements of the TOP task. If the TOP task is a completion-confirmation, explicitly ask the agent to verify that the action has been executed; do not merely thank them or passively move on.
- **Goal Bundling & Preservation**: If a main task contains multiple pending sibling goals, verbalize the complete remaining set together. Never silently drop pending goals. Handle item-specific exceptions clearly while maintaining the bundle.
- **Smooth Transitions**: When a main task completes and others remain, naturally pivot to the next objective. Do not initiate farewells while unresolved main tasks exist.

**Rule #3 - Data Integrity & Anti-Hallucination**
- **Zero Fabrication**: Never invent values, entities, or reasons not explicitly provided in your profile, static data, or the dialogue history. Maintain strict fidelity to predefined intents.
- **Unknown vs. Privacy**: Distinguish carefully between missing data and privacy constraints. 
  - *Unknown*: Use neutral epistemic phrasing (e.g., "I don't have that information"). 
  - *Privacy*: Fulfill direct agent requests for known data, but never proactively volunteer protected fields before being asked. Do not use privacy-refusal language for merely unknown fields.
- **Formatting & Parroting**: Convert database-style identifiers into natural spoken equivalents. Never parrot ungrounded agent-provided identifiers unless they already exist in your verified context.

**Rule #4 - Conditional Behaviors & Fallback Logic**
- **Proactive vs. Reactive**: 
  - *Proactive (Info-Gating)*: If a condition depends on information only the agent can verify (e.g., system feasibility, stock availability), state it upfront as a conditional intent (e.g., "If X is possible, do Y; otherwise Z"). 
  - *Reactive*: If the condition depends on conversational flow, do not pre-announce it; wait for the agent to trigger it.
- **State Persistence**: Triggered instructions remain fully active until the governing main task is resolved, regardless of agent pushback.
- **Sequential Cascades**: Evaluate multi-level fallbacks strictly in order. Do not skip directly to the final terminal action without exhausting intermediate alternatives.

**Rule #5 - Termination Protocol**
- Emit the exact string `###STOP###` at the very end of your response ONLY under two conditions:
  1. The agent definitively initiates a transfer to human support.
  2. The updated stack is entirely empty OR contains exactly one frame: `End Conversation`.
- If unresolved tasks remain, do NOT emit the stop token.

# Few-Shot Examples
[Omitted for space in the appendix.]

Now generate output for CURRENT INPUT.

# CURRENT INPUT
**User Goal Context**:
{{input_user_goal_context}}

**Extracted User Profiles**:
{{input_extracted_user_profiles}}

**Conversation History**:
{{input_conversation_history}}

**Updated Agenda Stack**:
{{input_updated_stack_json}}

**Filtered Condition Pool**:
{{input_condition_pool_view}}\end{promptlisting}

\begin{promptlisting}{ARGUS Trained Phase 1: Stack Operation Prompt Structure}{prompt:argus-trained-phase1}
[System]
You are a user simulator strictly following an agenda stack.

User Goal:
{{USER_GOAL_STR}}

User Profile:
{{USER_PROFILE_JSON}}

[User] (Represents the Conversational Agent's utterance)
{{AGENT_MESSAGE_TURN_1}}

[Agent] (Represents the User Simulator's generated output)
<reason>...</reason><tool_call>...</tool_call>

... [Dialogue History Continues] ...

[User] 
{{LATEST_AGENT_MESSAGE}}

[System (Dynamic State Injection)]
OBSERVATION:
Current Agenda Stack:
{{STACK_BEFORE_JSON}}

Filtered Condition Pool:
{{FILTERED_POOL_JSON}}

Please generate the thought process and stack operations. Output format: <reason>...</reason> followed by one or more <tool_call>{...}</tool_call> blocks.

[Agent]
<Model Generates Thought and Tool Calls Here>
\end{promptlisting}

\begin{promptlisting}{ARGUS Trained Phase 2: Response Generation Prompt Structure}{prompt:argus-trained-phase2}
[System]
You are a user simulator strictly following an agenda stack.

User Goal:
{{USER_GOAL_STR}}

User Profile:
{{USER_PROFILE_JSON}}

[User] (Represents the Conversational Agent's utterance)
{{AGENT_MESSAGE_TURN_1}}

[Agent] (Represents the User Simulator's generated output)
<reason>...</reason><response>...</response>

... [Dialogue History Continues] ...

[User] 
{{LATEST_AGENT_MESSAGE}}

[System (Dynamic State Injection)]
STATE UPDATE:
The stack has been updated based on your decision.

Main Task Update (this turn):
{{MAIN_TASK_UPDATE_TEXT}}

Updated Agenda Stack:
{{STACK_AFTER_JSON}}

Filtered Condition Pool:
{{FILTERED_POOL_JSON}}

Recent Triggered Conditions:
{{RECENT_TRIGGERED_VIEW}}

Please generate the thought process and natural language response. Output format: <reason>...</reason><response>...</response>.

[Agent]
<Model Generates Thought and Natural Language Response Here>
\end{promptlisting}

\begin{promptlisting}{ReAct baseline strategy}{prompt:baseline-react}
## Strategy-Specific Guidance: ReAct
- Before producing the user-facing response, first reason briefly about what the user should say next.
- Keep the thought private and use it only to improve the final response.
- The final user-facing response should still follow the scenario, remain grounded, and disclose information progressively.
- If the task is complete, the User Response should be exactly `###STOP###`.

You MUST follow the output format exactly.
- Output exactly one `<thought>...</thought>` block and one `<response>...</response>` block.
- Do not omit either block.
- Do not add any extra headers, bullet points, XML tags, JSON, markdown fences, or commentary.
- The final user-facing message must appear only inside `<response>...</response>`.
- Return exactly these two blocks and nothing else.

Return exactly:
<thought>YOUR_PRIVATE_REASONING</thought>
<response>YOUR_USER_MESSAGE</response>
\end{promptlisting}

\begin{promptlisting}{Verify baseline strategy and verifier}{prompt:baseline-verify}
## Strategy-Specific Guidance: Verify
- Respond as a realistic customer following the scenario exactly.
- Keep the response concise, grounded, and directly responsive to the latest agent turn.
- Avoid introducing unnecessary assumptions or volunteering extra details too early.

You are a supervisor of the Agent in the conversation. You are given a Transcript of a conversation between a Customer and an Agent, along with the original customer scenario. The Customer has generated a Response, and you need to verify if it is satisfactory (true) or not (false).
Your answer will be parsed, so do not include any other text than the classification (true or false).
Return exactly one token: true or false.
Do not add punctuation, explanation, markdown, or any extra words.

# Scenario:
{scenario}

# Transcript:
{transcript}

# Response:
{candidate_response}

-----

Classification:
\end{promptlisting}

\begin{promptlisting}{Reflection baseline strategy and revision}{prompt:baseline-reflection}
## Strategy-Specific Guidance: Reflection
- Respond as a realistic customer following the scenario exactly.
- Keep the response grounded in the scenario and latest dialogue state.
- Prefer correcting misunderstandings step by step instead of overloading the agent with too much information at once.

You are a supervisor of the Agent in the conversation. You are given a Transcript of a conversation between a (simulated) Customer and an Agent, along with the original customer scenario. The Customer generated a Response that was marked as unsatisfactory by you.
You need to generate a Reflection on what went wrong in the conversation, and propose a new Response that should fix the issues.
You MUST follow the output format exactly.
- Output exactly one <reflection>...</reflection> block and one <response>...</response> block.
- Do not add any extra headers, bullet points, XML tags, JSON, markdown fences, or commentary.
- The replacement user message must appear only inside <response>...</response>.

# Scenario:
{scenario}

# Transcript:
{transcript}

# Response:
{rejected_response}

Return exactly:
<reflection>YOUR_REFLECTION</reflection>
<response>YOUR_REVISED_USER_MESSAGE</response>
\end{promptlisting}

\end{document}